%% file: main.tex
\documentclass[times,final]{elsarticle}

\usepackage[english]{babel}

\usepackage[letterpaper,top=2cm,bottom=2cm,left=3cm,right=3cm,marginparwidth=1.75cm]{geometry}

\usepackage{amsmath}
\usepackage{setspace}
\usepackage{mathalfa}
\usepackage{graphicx}
\usepackage{float}
\usepackage[colorlinks=true, allcolors=blue]{hyperref}
\usepackage[table,xcdraw]{xcolor}
\usepackage{caption}
\usepackage{subcaption}
\usepackage{tikz}
\usepackage[ruled,vlined]{algorithm2e}
\usepackage{lineno}

\definecolor{rev1}{RGB}{0, 0, 0}
\definecolor{rev2}{RGB}{0, 0, 0}

\begin{document}

\begin{frontmatter}
\title{Reinforcement learning to maximise wind turbine energy generation}
\author[1]{Daniel Soler}
\author[1]{Oscar Mariño}
\author[1]{David Huergo}
\author[1]{Mart\'in de Frutos}
\author[1,2]{Esteban Ferrer}

\cortext[cor1]{Corresponding author}
\ead{esteban.ferrer@upm.es}

\address[1]{ETSIAE-UPM-School of Aeronautics, Universidad Politécnica de Madrid, Plaza Cardenal Cisneros 3, E-28040 Madrid, Spain}
\address[2]{Center for Computational Simulation, Universidad Politécnica de Madrid, Campus de Montegancedo, Boadilla del Monte, 28660 Madrid, Spain}

\begin{keyword}
Wind turbine \sep Blade element momentum theory \sep Reinforcement learning \sep double deep Q-learning \sep Value iteration
\end{keyword}

\begin{abstract}
We propose a reinforcement learning strategy to control wind turbine energy generation by actively changing the rotor speed, the rotor yaw angle and the blade pitch angle. A double deep Q-learning with a prioritized experience replay agent is coupled with a blade element momentum model and is trained to allow control for changing winds. 
The agent is trained to decide the best control (speed, yaw, pitch) for simple steady winds and is subsequently challenged with real dynamic turbulent winds, showing good performance. \textcolor{rev1}{The double deep Q-learning is compared with a classic value iteration reinforcement learning control and both strategies outperform a classic PID control in all environments.} Furthermore, the reinforcement learning approach is well suited to changing environments including turbulent/gusty winds, showing great adaptability.  \textcolor{rev2}{Finally, we compare all control strategies with real winds and compute the annual energy production. In this case, the double deep Q-learning algorithm also outperforms classic methodologies.}

\end{abstract}

\end{frontmatter}


\onehalfspacing

\section{Introduction}

Wind turbines play a crucial role in the generation of clean and renewable energy. However, effective control strategies are necessary to optimize their performance when faced with variable wind conditions.

Wind turbine control systems \cite{NJIRI2016377,NOVAESMENEZES2018945,https://doi.org/10.1049/rpg2.12160} aim to maximize energy generation while maintaining structural integrity and safe operating conditions. Furthermore, wind turbines operate in rapidly changing winds, and consequently, dynamic control strategies that adapt in variable environments are necessary.  Classic control mechanisms include, among others, modifying the yaw angle to optimize the orientation of the rotor and adjusting the rotational speed while controlling the pitch angle on the blades. 
These control approaches leverage real-time wind measurements, turbine dynamics, and advanced control algorithms to dynamically adjust yaw, rotor speed, and pitch simultaneously, allowing a more effective adaptation to changing wind conditions. This strategy leads to improved energy generation, reduced fatigue loads, and enhanced turbine longevity. 

The emergence of reinforcement learning (RL) offers new possibilities for wind turbine control by enabling data-driven adaptive decision making \cite{LECLAINCHE2023108354,GARNIER2021104973}.
RL is a machine learning approach in which an agent learns to make decisions in an environment to maximize cumulative rewards over time \cite{sutton1998rli}. In this paradigm, the agent explores the environment through interactions and takes actions to achieve a specific goal (e.g. the instantaneous optimal yaw, pitch, and rotational speed). The RL process is based on a feedback system in which the agent receives a reward or penalty signal after each action taken. Its goal is to learn a policy, which is the strategy that determines what action to take for each state of the environment in a way that maximizes the cumulative long-term reward. 

When applied to wind turbines, RL provides a way to autonomously learn how to adjust the control inputs to maximize power generation. RL algorithms capture complex nonlinear relationships between wind conditions, turbine states, and actions, continuously updating the policy based on the feedback received from the environment.
RL-based control methods are advantageous, as they can adapt in real-time to changing wind conditions. 
For these reasons, the integration of RL with wind turbine control has garnered significant attention from researchers due to its potential advantages. Several studies have already made substantial contributions in this area. In particular, previous works \cite{SESMPR20,SEAM20,SESM20,XIE2023118893,chen2020reinforcement} have proposed RL algorithms to control the pitch angle of a turbine, giving comprehensive information on reward definitions.
Additionally, \cite{KADOCHE2023119129,PARJ23} presented similar RL algorithms for yaw control. Moreover, previous studies by \cite{kushwaha2020q,wei2016adaptive} have explored rotor speed control for MPPT (maximum power point tracking) using Q-Learning, a specific RL algorithm. 
The studies presented in \cite{SAZEFULJLJ19,2SAZEFULJLJ19} demonstrated how RL systems can adapt to realistic wind conditions, enhancing overall energy generation at a targeted wind site. Furthermore, some authors have focused on devising control strategies for the yaw angle of individual turbines within a wind farm, with the aim of optimizing power generation while ensuring the safety of the wind turbine. For example, \cite{DHZJZX21} proposes an RL-based method for wake steering control, effectively optimizing wind farm efficiency. 

In this work, we explore the application of RL techniques for wind turbine control, including at the same time yaw, pitch, and rotational speed. Unlike the majority of previous studies, here we do not use an analytical model of wind turbine power but instead employ a more realistic Blade Element Momentum Theory (BEMT) model. This method is fast enough to be used in a reinforcement learning framework and allows the use of a wider control parameter space, higher flexibility in the turbine geometry and in the inflow wind conditions.
Additionally, we investigated the generalization of RL controllers in different operating conditions and validating winds. Finally, we compare two Reinforcement learning strategies to a classic PID control, showing the superiority of the proposed methodology.
By harnessing the adaptability and learning capabilities, our research aims to improve wind turbine control, contributing to more efficient and sustainable wind power generation.

The paper is organized as follows. First, we summarize the methodology in Section \ref{sec:method}, where we include the wind turbine model and the reinforcement learning strategy. We provide details on the reward, the neural network architecture, the training, and the validation. Second, in Section \ref{sec:results} we challenge the method with real winds and compare the RL control to a classic PID control. We end with conclusions and outlooks.

\section{Methodology}\label{sec:method}
In this section, we provide a brief description of the blade element momentum theory used to model the wind turbine, then detail the RL strategy.
\subsection{Wind turbine modeling using blade element momentum theory}
We select a wind turbine solver based on Blade Element Momentum Theory (BEMT) to model horizontal axis wind turbines \cite{Glauert1935,burton2011wind,https://doi.org/10.1002/we.1636}. 
 BEMT is very efficient and provides a simplified yet effective approach to estimate the aerodynamic forces and energy generation of wind turbines, allowing for fast function evaluations. This evaluation speed is key to training and validating the agent in a reasonable amount of time. Furthermore, numerical actuator discs, which are at BEMT's core, are used in complex wind farm simulations, including turbulent winds (e.g. large eddy simulations), see the review \cite{en14133745}.

In BEMT theory, the wind turbine blade is divided into small sections along its span, and the aerodynamic forces acting on each section are calculated based on the local wind conditions and the geometry of the airfoil. The local flow conditions, defined for each section and every time step, include the speed and direction of the wind and the turbulence intensity.
The aerodynamic forces, the lift and drag coefficients, are calculated using Xfoil \cite{drela1989xfoil}. The overall power and thrust generated by the wind turbine can be obtained by integrating the forces along the blades' span.

In this work, we use WISDEM's CCBlade \cite{NW19} for BEMT computations since it allows one to control the performance of the wind turbine through the set of parameters selected for this study: yaw, pitch, and rotational speed. 
The selected wind turbine is a large onshore wind turbine based on a SWT2.3-93, featuring a rated power of 2.3 MW. The turbine has been experimentally tested, and detailed geometry and data for benchmarking can be found in \cite{christophe_julien_2022_7323750,Churchfield,Nyborg}.

\subsection{Reinforcement learning strategy for wind turbine control}\label{subsec:RL_strategy}
In Reinforcement Learning, the agent interacts with the environment by performing actions that modify its current state. In response to these actions, the agent receives a reward from the environment, which allows the agent to learn whether the action was correct.

\begin{figure}[H]
  \centering
  \includegraphics[scale=0.55]{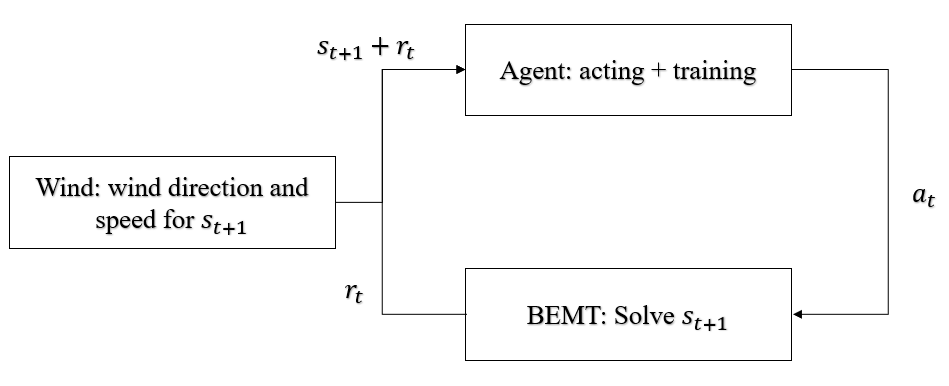}
  \captionsetup{labelfont={bf}}
  \caption{Flow diagram for the complete algorithm.}
  \label{fig:MDP }
\end{figure}

RL requires defining the state and action spaces, denoted by ${S}$ and ${A}$, respectively. However, the key point lies in the definition of the reward function, as it provides the feedback that will allow the agent to learn the best policy and optimize its behavior. By doing so, the agent learns how to act on the succession of states $s_i$, using actions $a_i$ and receiving rewards $r_i$ as feedback.
We summarize the algorithm, which resembles a Markov Decision Process, in Figure \ref{fig:MDP }, where the agent acts on the environment to control its parameters as a response to wind changes and receives rewards $r_t$ as feedback.

In our work, the goal is to teach the RL agent how to leverage past experience to act upon three control variables or degrees of freedom (DOFs): the yaw angle of the wind turbine, the pitch angle of the blades, and the rotor speed. The agent must learn to adapt to changing winds by varying the values of the control variables in order to boost energy generation.

The state of the agent must include all the information about the environment that is required to take the best action at each time step. If the state lacks some relevant information, the agent might not be able to achieve optimal behavior.
The state is defined by the value of the yaw misalignment, defined as the difference between the yaw and the wind direction in a fixed reference frame, the pitch angle, the rotor speed, and the wind speed and direction. However, not all possible combinations are allowed. These devices have mechanical limitations that constrain these variables. Furthermore, in the case of the rotor speed, there is a lower bound on the rotational speed (rpms) due to operational constraints. Consequently, we constrain our model to the following allowed values, which define the state space $S$:
\begin{itemize}
  \item yaw misalignment $\in [-30,30]$ degrees
  \item pitch $\in [-20, 20]$ degrees
  \item rotor speed $\in [6, 25]$ rpm
  \item wind speed $\in [4,13]$ m/s 
\end{itemize}
\textcolor{rev1}{
To maintain the validity of the airfoil and Blade Element Momentum Theory (BEMT) models, two additional constraints are imposed on the turbine operation. Specifically, for the airfoil model, it is imperative to ensure that the angle of attack of each blade (AoA) falls within a defined range. On the other hand, BEMT requires that the tip speed ratio (TSR), a dimensionless ratio of the wind turbine given by $\text{TSR}=\Omega R / U_\infty$, remain within certain bounds. Here, $\Omega$ represents the rotor speed in radians per second, $R$ denotes the wind turbine radius, and $U_\infty$ is the wind speed. These constraints are outlined below:}
\begin{itemize}
  \item TSR $\in [3, 12]$  
  \item AoA $\in [-10, 15]$ degrees
 \end{itemize}
\textcolor{rev2}{During the training phase, the agent must learn that these limitations exist by receiving a large punishment when trying to surpass those boundaries. In such instances, the agent is punished and the forbidden action is revoked so ${s}_{t+1} = {s}_t$. This approach has demonstrated to be the most effective option to make the agent learn the constraints of ${S}$. Alternative approaches, such as employing logarithmic or polynomial barrier functions on the reward, were explored, but proved to be less effective. Nevertheless, during the validation of the agent, it is allowed to execute any action, operating under the assumption that it has successfully learned the boundaries of the state space.}

The action space is defined by the DOFs considered in the problem. The turbine is controlled through the blade pitch angle, the rotor yaw angle, and the rotor speed. Consequently, each action is associated with the increase or decrease of each parameter. Therefore, our agent must learn to correctly assess every $a_i\in A$ as:\\
\begin{itemize}
\begin{spacing}{0.5}
  \item ${a}_{1}$: Increase yaw by one degree.
  \item ${a}_{2}$: Decrease yaw by one degree.
  \item ${a}_{3}$: Increase pitch by one degree.
  \item ${a}_{4}$: Decrease pitch by one degree.
    \item ${a}_{5}$: Increase rotor speed by one rpm.
  \item ${a}_{6}$: Decrease rotor speed by one rpm.
  \item ${a}_{7}$: Do not alter any of the DOFs.
  \end{spacing}
\end{itemize}
With this set of actions, the agent is able to navigate the state space $S$. The trained agent will be able to act on its three DOFs through 7 different actions that it must learn to evaluate for any $(s,a)$ combination. This will allow the agent to control the turbine to work at its optimal efficiency. 

\textcolor{rev1}{Once trained, the resulting agent should be able to adapt to any scenario, defined by its environmental conditions, while those conditions can be modeled through the variables defined in the state space, $S$, and the wind turbine can be controlled through the actions available in the action space, $A$.} 

\textcolor{rev2}{
In this work, we consider horizontal axis turbines, and define the state-action for this type of turbine. The  framework can potentially be generalized to other horizontal axis wind turbines (e.g. different rated power, blade geometries, rotor sizes, etc.) but the agents would require re-training. Note that the state-action space would need to be redefine to tackle  turbines that operate differently, such as vertical axis wind turbines. In the latter case, the yaw angle is not well defined and should not be considered for the state-action tuple.}

\subsubsection{The Deep Q-Network}
Q-learning is a popular reinforcement learning algorithm \cite{watkins1992q}. It belongs to the class of model-free RL algorithms, which means that it does not require prior knowledge or explicit models representing the system dynamics.
At the core of Q-learning is the Q-value, which represents the expected cumulative reward for taking a specific action in a given state. The Q-value is updated iteratively via the Bellman equation, which expresses the optimal action-value function in terms of the maximum expected future reward.

During the learning process, the wind turbine interacts with the environment, transitions between states, and takes actions according to its current policy. The Q-learning algorithm uses an \textit{epsilon greedy} exploration-exploitation trade-off to balance between exploring new actions and exploiting current knowledge to maximize cumulative rewards.
Initially, the Q-values are arbitrarily initialized and, as the wind turbine explores the environment and receives feedback in the form of rewards, the Q-values are updated using the temporal difference error. The temporal difference error represents the difference between the observed reward and the predicted reward based on the Q-values.
Through repeated iterations, Q-learning gradually converges to an optimal policy, where the wind turbine learns the best actions to take in different states, maximizing the power generation. 

\textcolor{black}{The Deep Q-Network (DQN) is a variant of Q-learning that uses a deep neural network to estimate Q-values \cite{mnih2013playing}. It replaces the traditional lookup table with a neural network, and allows generalizations across similar states to handle large state spaces efficiently. 
To train the DQN, an experience replay buffer is used. During the training phase, the agent interacts with the environment and stores the experiences (state, action, reward, next state) in the replay buffer. Then, random batches of experiences are sampled from the replay buffer to train the network and update its weights, helping to break the correlation between consecutive samples and improving stability during the learning process.}

\textcolor{black}{In this work, we go one step further, and the vanilla DQN is expanded with a Double Deep Q-Network (DDQN) \cite{van2016deep} and Prioritized Experience Replay (PER) \cite{schaul2015prioritized} algorithms.
The key innovation of DDQN is the use of two NNs: the main network and a target network. The target network is a separate neural network, with the same architecture as the main network, but with frozen parameters. 
The target network parameters are periodically updated to match the main network, allowing the target Q-Values to be changed slowly and preventing harmful feedback loops while minimizing biases.} 

\textcolor{black}{The PER algorithm samples the experiences, from the memory, for training according to a metric, as opposed to regular DQN random sampling. This metric is called ``priority'' and grows with the agent's error when calculating the associated Q-Value of a state-action $(s,a)$ transition. A very poorly predicted Q-value, that is, a state whose transition to the next state is yet unknown, will be much more likely to be sampled during the training than a low-priority transition (with which the agent is already familiar). 
By doing so, the agent is more likely to be trained with the experiences that it did not correctly predict. Note that these experiences are the most relevant for learning, resulting in a more efficient algorithm.}

\textcolor{rev2}{Although DQN has several advantages, the main drawback of Q-learning-based algorithms is that they are not suitable for a continuous space state. In this case, policy-gradient methods, such as PPO (Proximal Policy Optimization), provide a better approach for continuous states. When controlling wind turbines, the yaw and pitch angles and the rotor speed can be discretized while keeping the size of the state space small enough. Furthermore, real turbines require actuators to change these control parameters, which typically perform discrete actions. Therefore, the use of DQN seems justified for this application. In general, DQN requires less training time than PPO and is less prone to converge to local optima. For these reasons, we will use DQN as the main algorithm to test the proposed methodology.}

\textcolor{rev1}{Additionally, we include comparisons with a different RL strategy, based on Value Iteration (VI), which is detailed in \ref{appendix_Value_Iteration}. VI is considered a classic RL algorithm \cite{sutton1998rli} that does not require a neural network; instead, it is a tabular-based method. The main advantage of this algorithm is that convergence is mathematically guaranteed. Since it relies on tabulated data, it can only be used if the state space is small. Therefore, if we can prove that our DQN approach, which is highly scalable and flexible, shows a behavior comparable to that of the VI agent for wind turbine control, then we will have an accurate DQN agent that is widely generalizable for complex problems.}

In the context of wind turbine control, we have adapted Q-learning to optimize pitch angle, yaw angle, and rotor speed. By leveraging the learning capabilities of Q-learning, wind turbines can adapt their control strategies in real time, leading to improved power production capabilities and enhanced overall performance.
In the following sections, we will dive into the specific application of Q-learning in wind turbine control and discuss its performance and challenges.

\subsubsection{Neural network architecture}
\color{rev2}
When using DQN, neural networks (NN) approximate the Q-Function. The NN receives various input parameters to capture the state-action pairs: yaw misalignment $\tilde\gamma$, defined as the difference between yaw and wind direction; pitch angle $\theta$; rotor speed $\Omega$; wind speed $U_\infty$; and a binary-encoded index representing the last chosen action. The NN outputs the Q-Value for the given state-action pair $(s,a)$. This process is represented by the following mapping:
$$Q(s,a)\approx\mathcal N [\underbrace{U_\infty,\tilde\gamma,\Omega,\theta}_{\textrm{state}},a;W].$$

\begin{figure}[htbp]
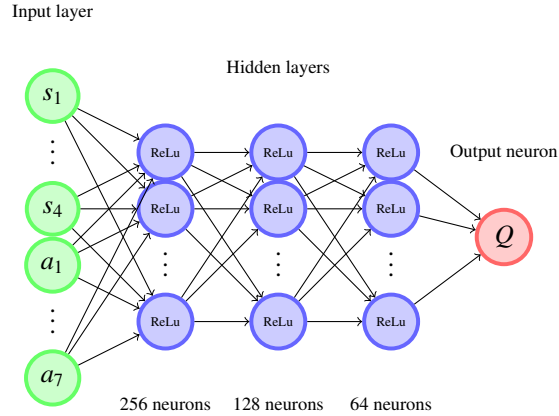

    \centering
    \include{NN_scheme}
    \captionsetup{labelfont={bf}}
    \caption{Neural network architecture.}
    \label{fig:NN_scheme}
\end{figure}

Here, $W$ denotes the weights and biases of the NN. The neural network architecture adopts a Multi-Layer Perceptron (MLP) structure, formed by three dense hidden layers with Rectified Linear Unit (ReLU) activation functions. The last layer, which is only the output neuron, has a linear activation function instead of a ReLU. This is done to allow the Q-values to have any sign and not just positive numbers. Figure \ref{fig:NN_scheme} visually illustrates this architecture. The first layer uses 256 neurons followed by layers with 128 and 64 neurons, as shown in the figure.

\color{black}
\color{rev2}

The loss function used for training in the Double Deep Q-Network (DDQN) algorithm follows the mean squared error criterion, comparing the predicted Q-values from the online network, $Q(s,a;W)$, with the target Q-values, $\tilde Q(s,a;\tilde W)$, as derived from the Bellman equation:
$$\mathcal L = \mathbb E[(r(s,a)+\gamma \max_{a'} \tilde Q(s',a';\tilde W)-Q(s,a;W))^2]+\lambda\lVert W\rVert_2^2,$$\\
where $r(s,a)$ is the reward function and $\gamma$ is the discount factor. Notice that a L2 regularization term has been incorporated into the loss function. This is done to enhance the generalization of the model and prevent overfitting. This regularization has a constant $\lambda$ associated. The weights of the target network, $\tilde W$, are updated at each training iteration using a soft update rule, governed by the hyperparameter $\tau_{soft \phantom{1} update}$:
$$ \tilde W = \tau_{soft \phantom{1} update}\cdot W + (1-\tau_{soft \phantom{1} update})\cdot\tilde W.$$

The neural network training is implemented through \textit{Keras} \cite{CF15}, providing a straightforward coding interface for optimization. In our approach, we employ the Adam optimizer \cite{kingma2014adam}, a widely recognized state-of-the-art stochastic gradient descent optimizer. As we have already mentioned, we have integrated the DDQN with a PER which is a prioritized sample strategy, utilizing an Unsorted Sum Tree \cite{J18} for its implementation.

\color{black}
The agent is evaluated using two sets of hyperparameters: an arbitrary and an optimized. The latter results arise from running a Bayesian optimization on the RL algorithm using the \textit{Optuna} library \cite{ATSSYYOYKM19,optuna_2019}.
The values for both sets of hyperparameters are detailed in Table \ref{tab:hyperparameters}.

\begin{table}[htbp]
\begin{tabular}{l||c|c}
\hline
\bf Hyperparameter       & \bf  Arbitrary Set & \bf Optimized Set \\ \hline
Learning rate        & 0.01          & 0.00033     \\ 
Episodes             & 250           & 149           \\ 
$\alpha_{PER}$       & 0.75          & 0.6711        \\ 
$\epsilon_{PER}$     & 0.01          & 0.01          \\ 
Batch size           & 32            & 16            \\ 
Epochs               & 3             & 3             \\ 
$\tau_{soft \phantom{1} update}$ & 0.1           & 0.1\\ 
$\epsilon_{greedy}$  & 0.2           & 0.3           \\ 
Discount Factor      & 0.95          & 0.95          \\ 
L2 strength          & 0.001         & 0.00859     \\\hline
\end{tabular}
\centering
\captionsetup{labelfont={bf}}
\caption{Hyperparameter sets for the agent.}\label{tab:hyperparameters}
\end{table}

\color{rev2}
The change in hyperparameter causes unequal resources consumption during training and validation. Longer training phases (250 episodes) are associated with the Arbitrary Set, which requires longer time and more memory than shorter ones, such as the Optimized Set's 149 episodes. This is due to the increased number of iterations needed for training. Additionally, Table \ref{tab:resources} shows how each combination of hyperparameters demands distinct resources for completing each episode: the optimized set of hyperparameters concludes each training episode in less time, thus demanding less memory. This behavior is derived from the fact that the optimized agent is capable of winning episodes both earlier and at a higher rate, thus demanding fewer iterations per episode. Therefore, the optimized agent is not only more potent, as will be discussed during the results section, but also more efficient because of both needing fewer iterations per episode and training for fewer episodes.
\color{black}

\begin{table}[htbp]
\centering
\begin{tabular}{l||c|c|c|c}
\hline
\bf Hyperparameter Set & \bf CPUs & \bf Average memory (MB/episode) & \bf Average time (s/episode) & \bf Episodes \\ \hline
Optimized & 1 & 569 & 126 & 149 \\
Arbitrary & 1 & 708 & 139 & 250 \\ \hline
\end{tabular}
\captionsetup{labelfont={bf}}
\caption{Resources needed for training and validation.}
\label{tab:resources}
\end{table}

\subsubsection{Reward definition}
The reward function is the key to learning for the agent, as it is the only feedback source on how successful are the selected actions. The reward must be carefully crafted for each specific problem to learn an appropriate policy. 
The ideal reward should guide the agent to obtain the maximum power production regardless of the wind conditions. 

For example, to teach the agent to maximize the power controlling only the yaw angle, there are two different options. First, the reward function could directly encourage the agent to align itself with the incoming wind. Alternatively, a reward function, that only rewards higher energy generation, could be implemented, and the agent should also learn to align itself with the wind. The latter is better since avoiding specific reward functions allows greater flexibility and stability during training. In addition, it eases the task of learning complex behaviors that depend on multiple control variables. Moreover, such an unbiased reward avoids local optima that may have been wrongfully identified as the global optimum during a preliminary study, since the agent will determine for itself such optimal working conditions. For all of these reasons, we choose the power coefficient ${C_p}$ of the turbine, as the reward because it only promotes maximum power generation and only depends on the working conditions of the turbine relative to the wind (rather than on the wind conditions). Therefore, the reward function is defined through the ${C_p}$ of the turbine, which is set to linearly grow from ${r} = 0$ when ${C_p} = 0$ to ${r} = 1$ when ${C_p} = {C}_{p,nom}$, where ${C}_{p,nom}$ is the maximum possible value of the power coefficient when the turbine is working under optimal conditions (nominal power) and $r$ stands for reward.

This power-based reward is complemented by two additional rewards. First, the agent receives punishments whenever it performs a forbidden action, that is, an action that leads to a state $s_{t+1}\notin S$. In such cases, the agent receives a negative reward with a value of $r = -2$ and the action is revoked so that $s_{t+1} = s_t$. Second, a large reward is provided when the agent achieves  97.5\% of $C_{p,nom}$ during 20 time steps in a row (i.e. the agent has won the game in this case). \textcolor{rev1}{Consequently, we can summarize the reward function as follows:}
\color{rev1}
\begin{equation*}
    r(s_t,a_t) =
    \begin{cases}
        \quad\displaystyle\frac{C_p}{C_{p,nom}}\\
         \qquad -2 & \text{if } s_{t+1}\notin S\\
         \qquad +5 & \text{if } \min\{r_i\}_{i=t-20}^t\ge0.975.
    \end{cases}
\end{equation*}
\color{black}

\textcolor{rev1}{Finally, it is important to notice that the reward depends on the selected wind turbine, whose specifications allow to obtain a target generated power. Therefore, once trained, the agent will be optimized for that specific turbine. The same agent could be used to control a different wind turbine, if the new turbine is similar to the one used as a reference, but of course the control might not be optimal. Hence, it is advisable to train a new agent for each specific application to maximize performance.}

\subsubsection{Wind turbine control and  expected results}\label{subsubsec:Turbine_control_expected_results}
Before using RL, we have performed a study of the possible states of the wind turbine, which will be used for understanding and validating the RL optimization. Indeed, having three adjustable parameters (pitch, yaw, and rotor speed) provides a variety of control strategies. 

\begin{figure}[H]
  \centering
  \begin{subfigure}{0.45\textwidth}
    \centering
    \includegraphics[width=\linewidth]{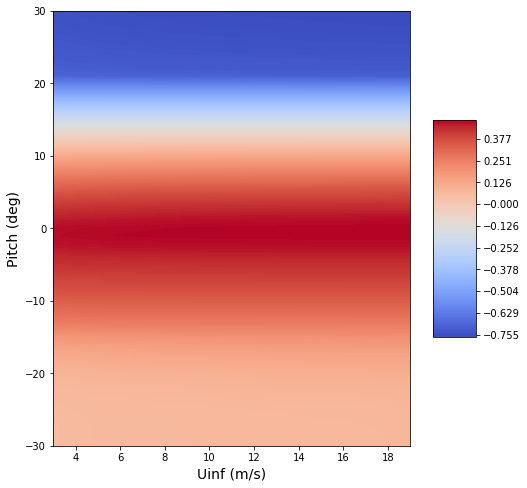}
    \captionsetup{labelfont={bf}}
    \caption{$C_p$ surface with fixed TSR of 8.}
    \label{fig:cp cut fixed tsr}
  \end{subfigure}
  \hfill
  \begin{subfigure}{0.45\textwidth}
    \centering
    \includegraphics[width=\linewidth]{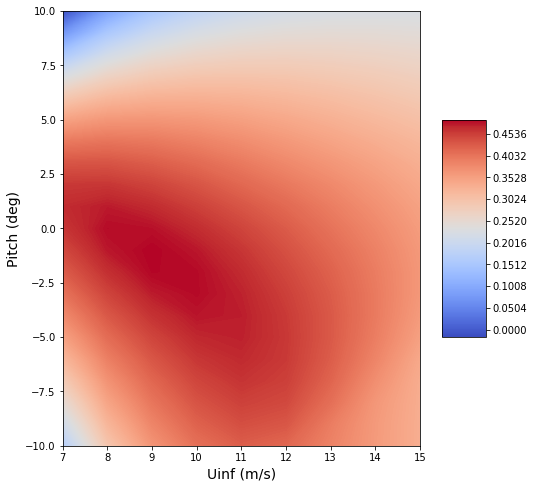}
    \captionsetup{labelfont={bf}}
    \caption{$C_p$ surface with fixed rotor speed.}
    \label{fig:cp cut fixed rpm}
  \end{subfigure}
  \captionsetup{labelfont={bf}}
  \caption{Power coefficient $C_p$ for: (a) a fixed tip speed ratio (TSR) of 8  and (b) a fixed rotational speed. }
  \label{fig:cp surfaces}
\end{figure}

Using blade element momentum theory, we can predict the power coefficient $C_p$ for a fixed tip speed ratio (TSR) of 8, shown in Figure \ref{fig:cp cut fixed tsr}, and for a fixed rotational speed, shown in Figure \ref{fig:cp cut fixed rpm}. 

In the left Figure \ref{fig:cp cut fixed tsr}, the control by maintaining a constant TSR scenario is presented. Following such a strategy, if the agent adjusts the rotor speed according to the wind speed, it can retain the TSR fixed, and acting on the pitch angle is not necessary. On the right Figure \ref{fig:cp cut fixed rpm}, we fix the rotor speed. Under these conditions, the optimum control could be achieved by adjusting the pitch angle and forgetting about the rotor speed. A third possibility would be to control both the pitch and the rotor speed at the same time.

%
The impact of the different control parameters on power generation is very different in each case. Varying the pitch can have around 37 times more impact on power generation than varying the yaw. Varying the rotor speed can have 12 times more impact than varying the yaw when the state is near the optimum conditions.

This preliminary study concludes that the pitch and rotor speed actions should receive larger rewards than the yaw actions, leading to prioritized control of the pitch and rotor speed over the yaw. 

\subsubsection{Training}

\begin{figure}[H]
  \centering
  \begin{subfigure}{\textwidth}
    \centering
    \includegraphics[width=1\linewidth]{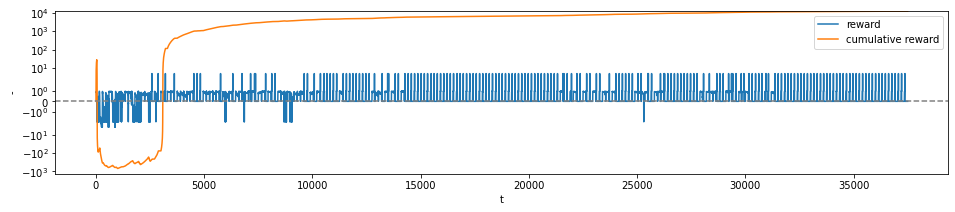}
    \captionsetup{labelfont={bf}}
    \caption{Instant and cumulative reward during training.}
    \label{fig:reward training}
  \end{subfigure}

  \vspace{1em} 

  \begin{subfigure}{\textwidth}
    \centering
    \includegraphics[width=1\linewidth]{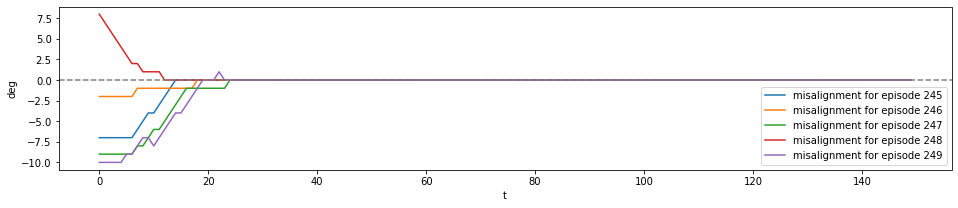}
    \captionsetup{labelfont={bf}}
    \caption{Yaw during the last episodes.}
    \label{fig:yaw training}
  \end{subfigure}

  \vspace{1em} 

  \begin{subfigure}{\textwidth}
    \centering
    \includegraphics[width=1\linewidth]{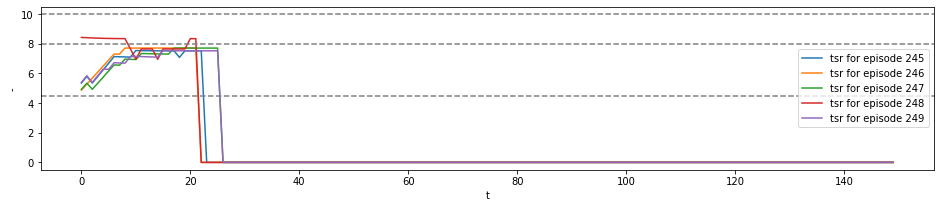}
    \captionsetup{labelfont={bf}}
    \caption{TSR during the last episodes.}
    \label{fig:tsr training}
  \end{subfigure}
  
  \captionsetup{labelfont={bf}}
  \caption{Wind turbine training metrics.}
  \label{fig:training}
\end{figure}

During the training phase, the agent faces random steady wind conditions during short episodes of up to 150 time steps. This allows the agent to adapt to virtually any wind. Even if the wind changes more rapidly than every 150 time steps, the agent is still capable of adapting. 
Furthermore, this training allows the agent to learn the limitations of the set of states $S$ and also to win all training episodes. To check the effectiveness of the learning progress during training, the cumulative reward is plotted against time in Figure \ref{fig:reward training}. \textcolor{rev2}{The figure illustrates the learning process for the agent. At the beginning ($t<800$), the agent receives negative rewards (blue curve) showing that it does not know the right actions. After this initial learning phase, the rewards become positive almost all the time, showing that the agent has learned how to maximize the rewards. We observe sporadically negative rewards (after the initial phase) linked to exploration, but overall the agent knows how to maximize the rewards (in this case the wind turbine energy).} 
Similarly, the effectiveness of the training process can be observed in the rapid growth of the cumulative reward (orange curve) and when checking the good performance of the agent during the last five training episodes. The associated yaw and TSR values are depicted in Figures \ref{fig:yaw training} and \ref{fig:tsr training}, respectively. 
First, Figure \ref{fig:yaw training} shows that the agent has learned to align the rotor perpendicular to the incoming wind. Second, Figure \ref{fig:tsr training} shows that the agent is capable of adequately tuning the turbine TSR to find the optimal value which, combined with the yaw control, leads to the maximum power generation. Moreover, both figures show how the agent has learned to prioritize TSR over yaw misalignment. These results agree with the expected results outlined in Section \ref{subsubsec:Turbine_control_expected_results}.

\subsubsection{Validation scenarios}
\begin{figure}[H]
  \centering
  \includegraphics[width=0.7\textwidth]{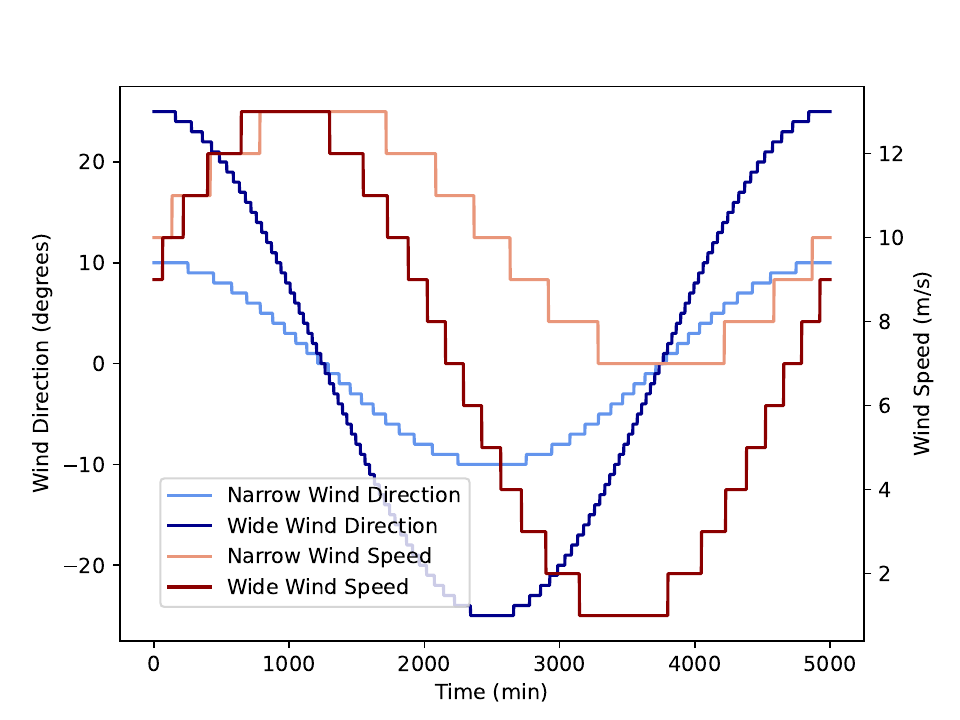}
  \captionsetup{labelfont={bf}}
  \caption{\textit{Narrow} and \textit{Wide} wind environments used for validation.}
  \label{fig:sin winds}
\end{figure}

\begin{figure}[h]
  \centering

  \begin{subfigure}{0.49\textwidth}
    \centering
    \includegraphics[width=\textwidth]{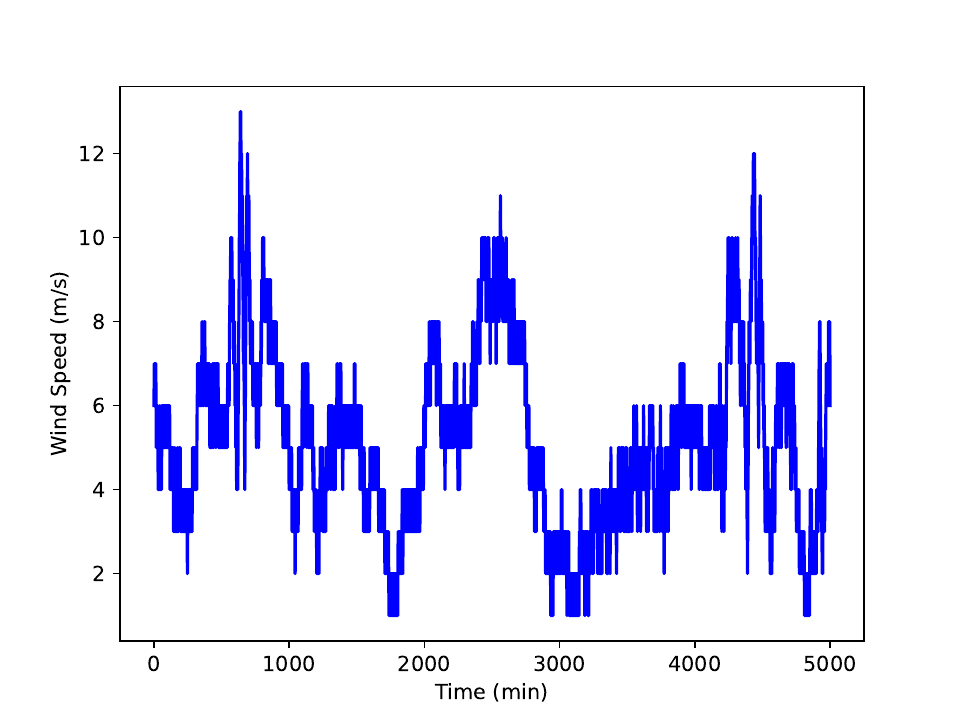}
    \captionsetup{labelfont={bf}}
    \caption{Wind Speed.}
    \label{fig:exp_wind_speed}
  \end{subfigure}
  \hfill
  \begin{subfigure}{0.49\textwidth}
    \centering
    \includegraphics[width=\textwidth]{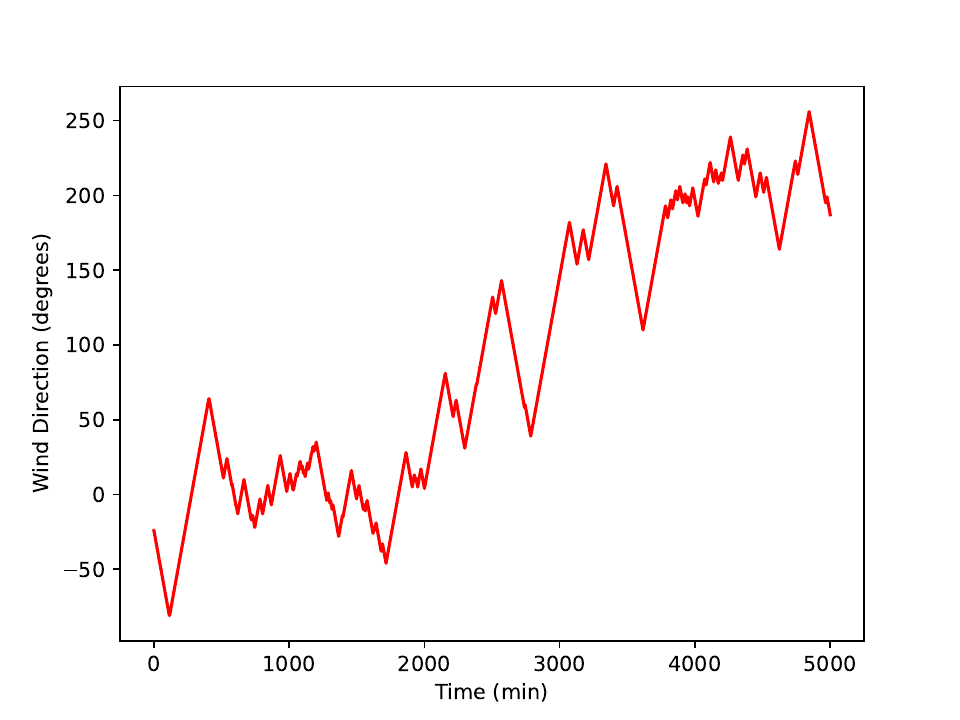}
    \captionsetup{labelfont={bf}}
    \caption{Wind direction.}
    \label{fig:exp_wind_dir}
  \end{subfigure}
  
  \captionsetup{labelfont={bf}}
  \caption{\textit{Experimental} wind used for validation.}
  \label{fig:experimental_wind}
\end{figure}

Once the agent has been trained, the agent is validated against three different winds to test its adaptability and performance. Each selected environment poses a different challenge: a wider allowed state-space $S$, a faster rate of change in wind conditions, or both. The simplest environment, called the \textit{narrow}, consists of sinusoidal variations of wind speed and direction in a small set $S$, where the wind speed ranges from 7 to 13 m/s and the wind direction ranges from -10 to 10 degrees (with respect to a fixed direction). The second environment, called the \textit{wide} environment, is the same as the \textit{narrow} one but with a wider set of states in $S$ with wind speeds ranging from 1 to 13 m/s in intensity and wind directions that range from -30 to +30 degrees in relation to the same fixed direction. 
Both environments are plotted in Figure \ref{fig:sin winds}, starting with random phases.
Finally, we also include an \textit{experimental} environment. This wind consists of real wind data measured by NREL's M2 tower \cite{JAA96} between the 10th and 14th of July 2023, which has been edited to be continuous and limited to change direction or intensity only once per time step. In this way, if the agent makes the correct decision (one decision per time step), the turbine should always operate under near-optimal conditions. These conditions are plotted in Figure \ref{fig:experimental_wind}. \textcolor{rev2}{ The figure shows that the wind is unsteady, turbulent and gusty showing rapid changes in time, and is therefore a challenging environment to test our reinforcement learning algorithm.}

\section{Results}\label{sec:results}
In this section, the performance of the RL agents is assessed in different environments. To do so, we consider five agents:
\begin{enumerate}
    \item DDQN agent with optimized hyperparameters (DDQN1), 
    \item DDQN agent with arbitrary hyperparameters  (DDQN2), 
    \item \textcolor{rev1}{Value Iteration agent (VI) (see \ref{appendix_Value_Iteration} for details),}
    \item Classic PID controller (see \ref{appendix_PID} for details),
    \item Fixed operating conditions (uncontrolled agent).
\end{enumerate}

\color{rev1}
To provide a broader context for the results, we will compare the DDQN agents with classical alternatives such as Proportional-Integral-Derivative (PID) control and a conventional RL algorithm, the Value Iteration (VI).

\color{rev2}
To measure the performance of the five agents, various metrics can be considered. One of the most popular ones for wind turbine farms is the capacity factor, $CF = \displaystyle\frac{E}{P_{nom}\cdot T}\cdot 100\%$, where $E$ is the energy produced during some period $T$ and $P_{nom}$ is the nominal power. However, it may not be the most suitable metric for this task due to its dependence on wind conditions. Thus, we introduce the Control Capacity Factor (CCF), defined as the ratio between the average power coefficient and the nominal or maximal power coefficient, i.e, $CCF = \displaystyle\frac{\langle C_p\rangle}{C_{p,nom}}\cdot 100\%$. This metric is designed to focus exclusively on the agent's ability to control the turbine, and independently of wind variations.
\color{black}

\begin{figure}[H]
  \centering
  \includegraphics[scale=0.55]{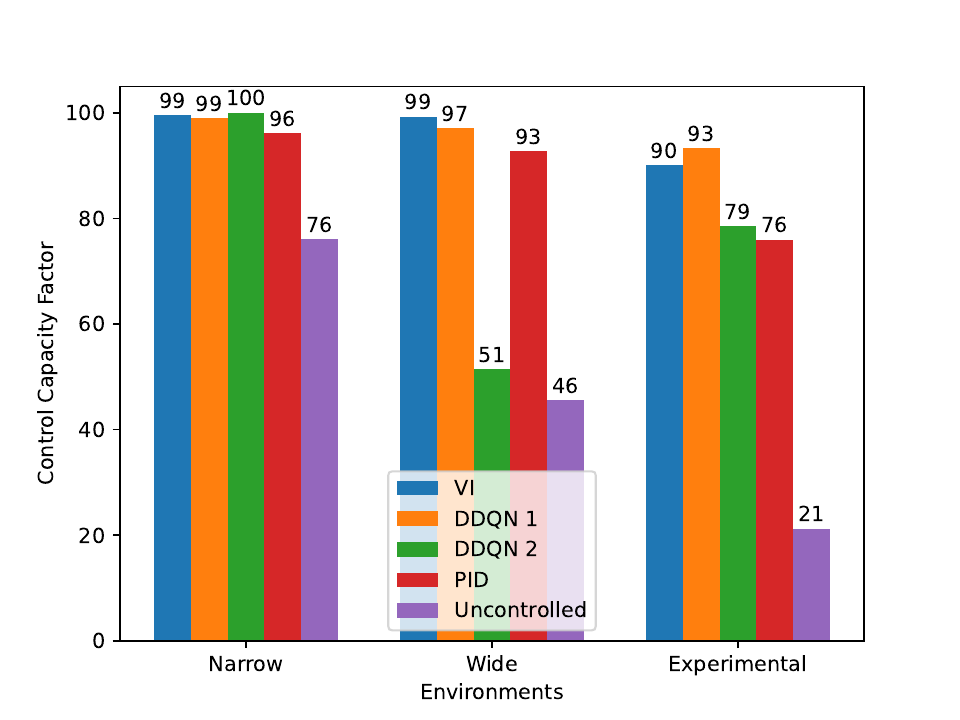}
  \captionsetup{labelfont={bf}}
  \caption{Performance of five control strategies across three wind environments.}
  \label{fig:performance review}
\end{figure}

\begin{figure}[H]
  \centering
  \begin{subfigure}{0.49\textwidth}
    \centering
    \includegraphics[width=0.9\linewidth]{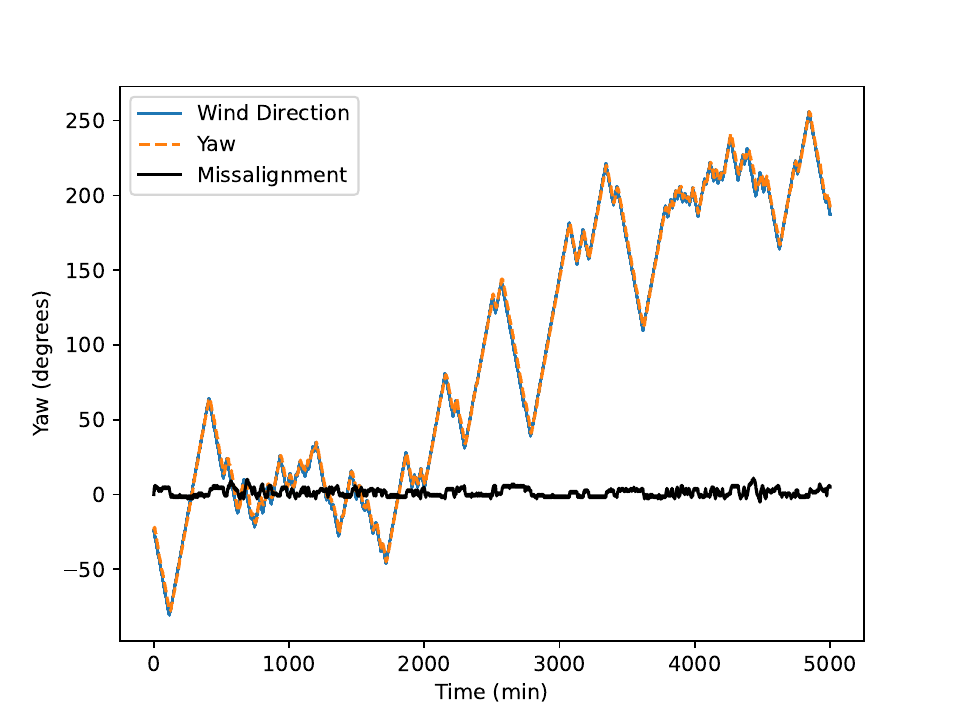}
    \captionsetup{labelfont={bf}}
    \caption{Yaw control plot.}
    \label{fig: yaw exp control}
  \end{subfigure}
  \hfill
  \begin{subfigure}{0.49\textwidth}
    \centering
    \includegraphics[width=0.9\linewidth]{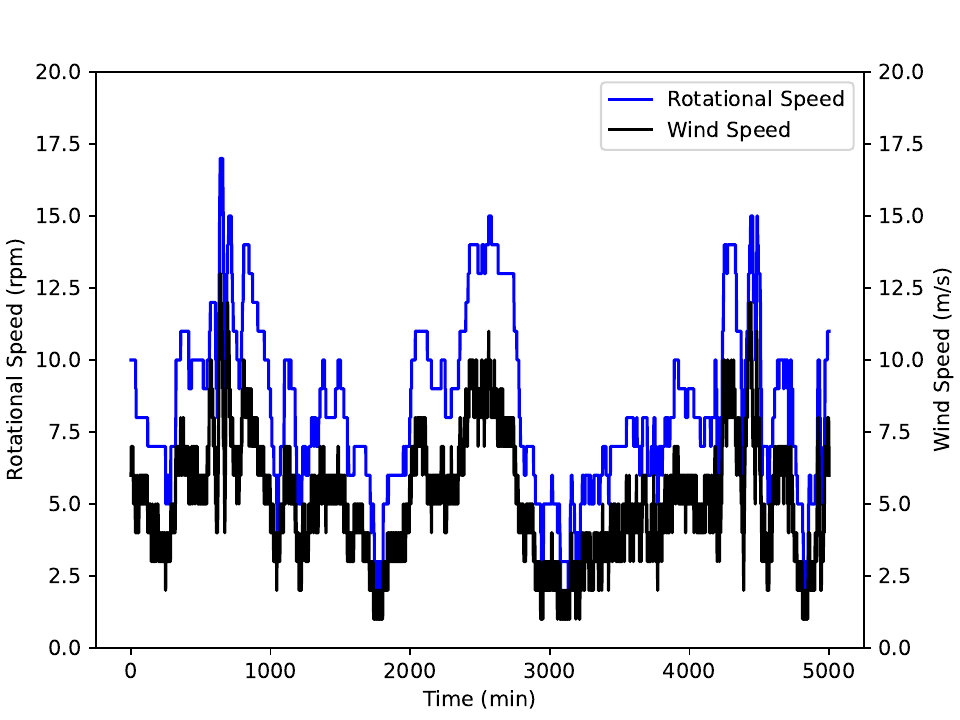}
    \captionsetup{labelfont={bf}}
    \caption{Rotor speed and wind speed control plot.}
    \label{fig:rpm exp control}
  \end{subfigure}

  \begin{subfigure}{0.49\textwidth}
    \centering
    \includegraphics[width=0.9\linewidth]{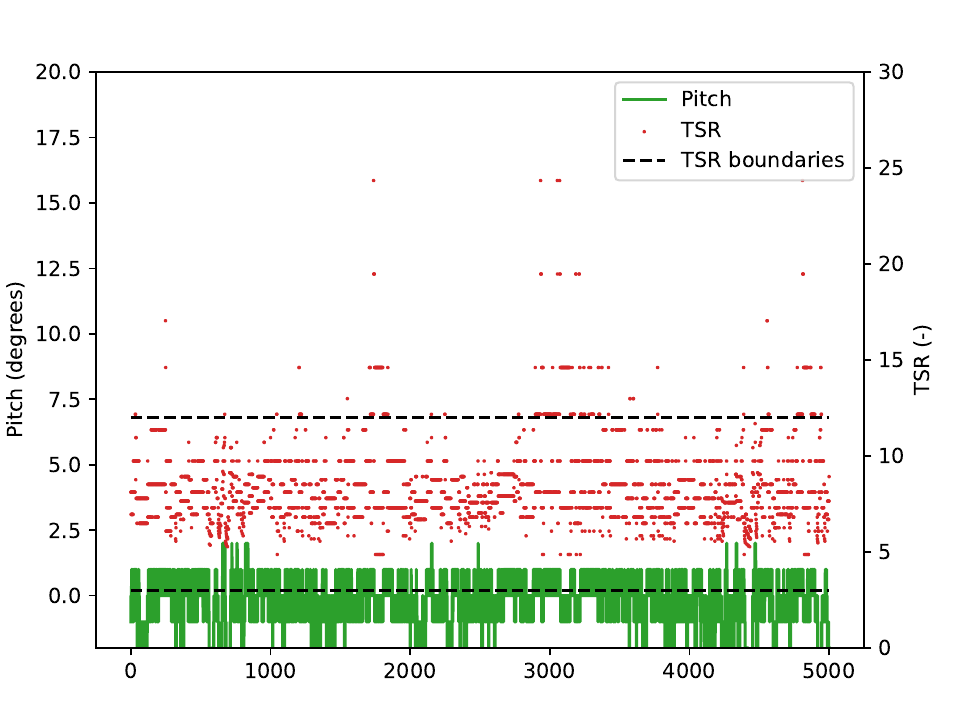}
    \captionsetup{labelfont={bf}}
    \caption{TSR evolution and pitch control plot.}
    \label{fig:tsr exp control}
  \end{subfigure}
  \hfill
  \begin{subfigure}{0.49\textwidth}
    \centering
    \includegraphics[width=0.9\linewidth]{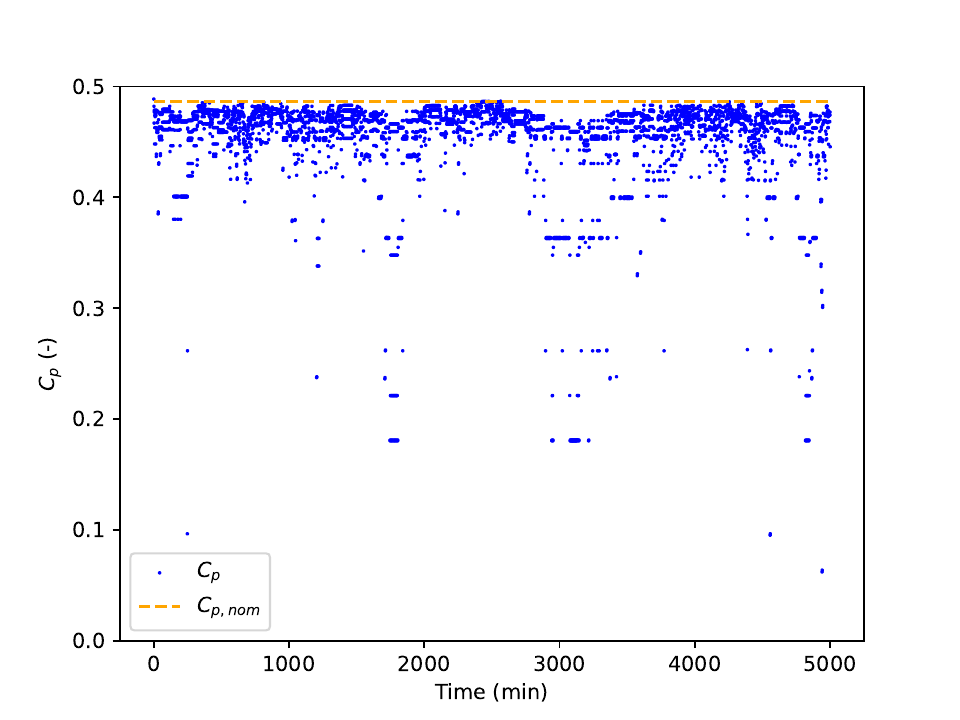}
    \captionsetup{labelfont={bf}}
    \caption{$C_p$ control plot.}
    \label{fig:cp exp control}
  \end{subfigure}

  \captionsetup{labelfont={bf}}
  \caption{Controlled variables for the DDQN1 agent on the \textit{experimental} environment.}
  \label{fig:DDQN exp control}
\end{figure}

The performance of each agent in the three validation environments is shown in Figure \ref{fig:performance review}.  In this plot, the DDQN1 agent (depicted in orange) and the Value Iteration agent (depicted in blue) are the most efficient for all environments. Their performance (although eclipsed in the \textit{narrow} environment by the perfect control achieved by the DDQN agent with arbitrary hyperparameters, DDQN2) is consistently better than the classic PID control and, of course, much better than the uncontrolled agent. Moreover, the optimized DDQN1 agent experiences a less severe drop in performance as environments become harder, proving its adaptability. The difference in performance between DDQN agents is remarkable and emphasizes the relevance of carefully tuning the agent's hyperparameters.
The optimized DDQN1 agent emerges as the best option for controlling the turbine under the \textit{experimental} environment. The results for this particular environment-agent combination are the most interesting and will be discussed in greater detail. In Figure \ref{fig:DDQN exp control}, the control variables for the best combination are shown during validation. Figure \ref{fig: yaw exp control} shows an effective yaw control that is capable of aligning the turbine (depicted in orange) with the wind (depicted in blue), achieving a very small misalignment of 2.22 degrees on average, which does not hinder power generation. Figures \ref{fig:rpm exp control} and \ref{fig:tsr exp control} show how the agent resorts to the use of both the pitch angle and the rotor speed to optimize the aerodynamic efficiency of the blades. Section 2.2.4 discussed how the pitch actions had a significantly greater impact on the $C_p$ than rpm actions, thus providing greater rewards. These rewards have taught the agent to control the aerodynamic efficiency primarily through the pitch angle, which is used at a rate four times larger than the rpms. The rpm actions have indeed lower effect on the $C_p$, which proves to be beneficial for fine control when near the optimum. The agent has learned how to perform a coarse control via pitch actions complemented with a finer control via rpm actions. This combined strategy allows the agent to maintain a very high CCF value, while coping with unseen winds.

Figure \ref{fig:tsr exp control} shows the TSR at each iteration. As shown in Figure \ref{fig:cp cut fixed rpm} the optimal TSR value is a function of the pitch angle which explains its fluctuations. \textcolor{rev2}{Note that there are instances where the tip speed ratio exceeds the specified boundaries. This occurs particularly during very low wind speeds, possibly due to a lower exposure to low-speed conditions during the training phase. This leads to a deviation in the power coefficient, causing it to drop, as observed in Figure \ref{fig:cp exp control}. However, in an actual wind turbine control, such events would not occur, as these low wind speeds fall below the cut-in speed for the wind turbine, triggering a shutdown}. The strategy found by the DDQN of controlling the yaw and TSR turns out to be very efficient to generate energy, reaching 93.3\% of the maximum attainable $C_p$, which is the $C_{p,nom}$, as shown in Figure \ref{fig:cp exp control}.

The agent, which has been trained on steady/stationary conditions, is remarkably able to adapt to an environment that is changing once per time step and is achieving a very high efficiency regardless of the wide state space and rapid changes. 
Furthermore, let us note that the RL agent has proven to be very flexible, since we used the exact same weights for all environments and achieved 99\% of CCF in the \textit{Narrow} environment and a 97\% of it on the \textit{Wide} one.
\textcolor{rev1}{These results show the ability of the trained RL agent to adapt to a wide variety of different conditions. Therefore, this agent can be used for any wind conditions if they can be completely defined by the wind speed and the wind direction.}

\begin{figure}[H]
  \centering
  \begin{subfigure}{0.49\textwidth}
    \centering
    \includegraphics[width=0.9\linewidth]{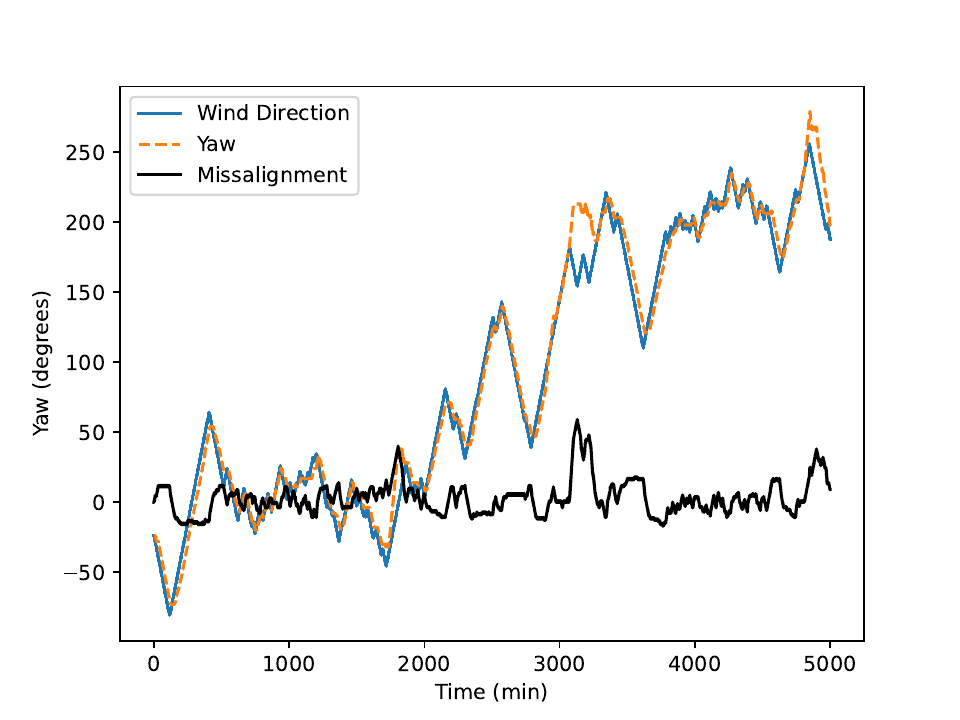}
    \captionsetup{labelfont={bf}}
    \caption{Yaw control plot.}
    \label{fig: VI yaw exp control}
  \end{subfigure}
  \hfill
  \begin{subfigure}{0.49\textwidth}
    \centering
    \includegraphics[width=0.9\linewidth]{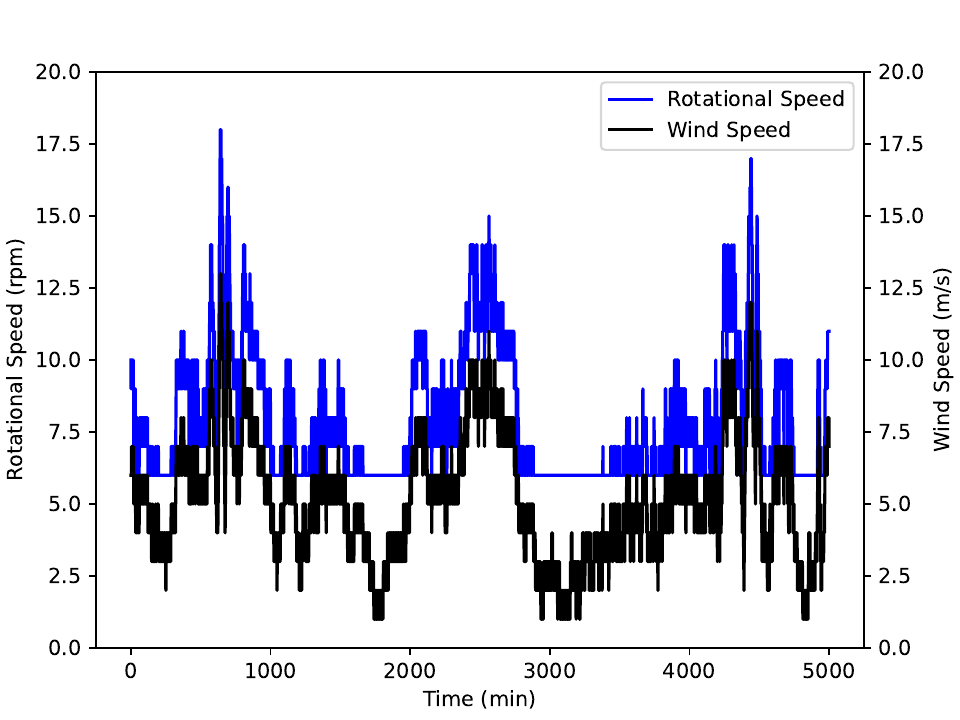}
    \captionsetup{labelfont={bf}}
    \caption{Rotor speed and wind speed control plot.}
    \label{fig: VI rpm exp control}
  \end{subfigure}

  \begin{subfigure}{0.49\textwidth}
    \centering
    \includegraphics[width=0.9\linewidth]{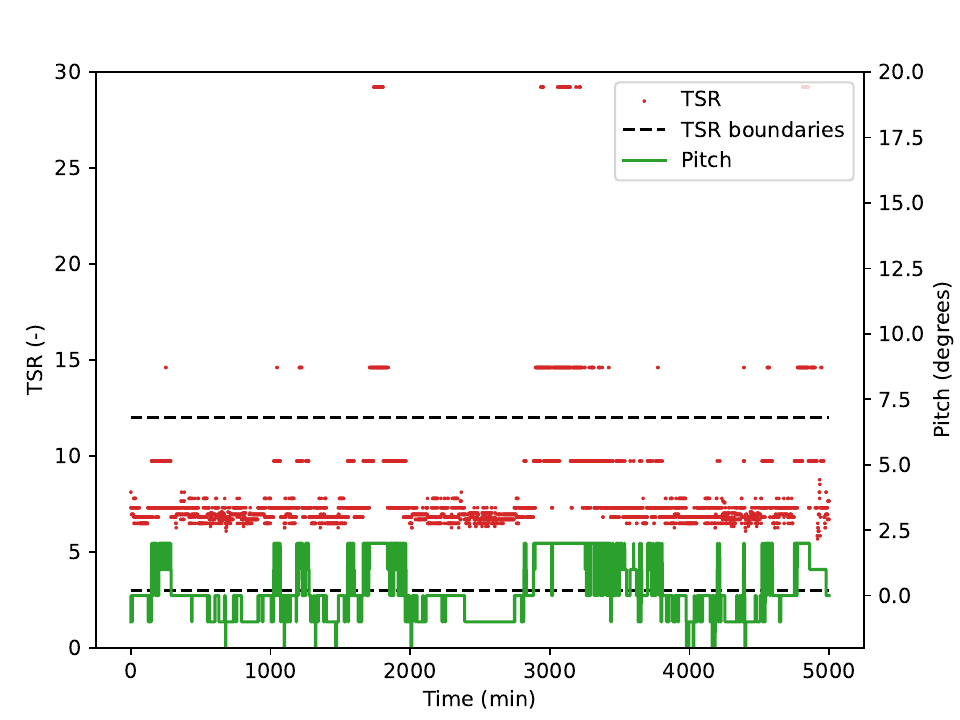}
    \captionsetup{labelfont={bf}}
    \caption{TSR evolution and pitch control plot.}
    \label{fig: VI tsr exp control}
  \end{subfigure}
  \hfill
  \begin{subfigure}{0.49\textwidth}
    \centering
    \includegraphics[width=0.9\linewidth]{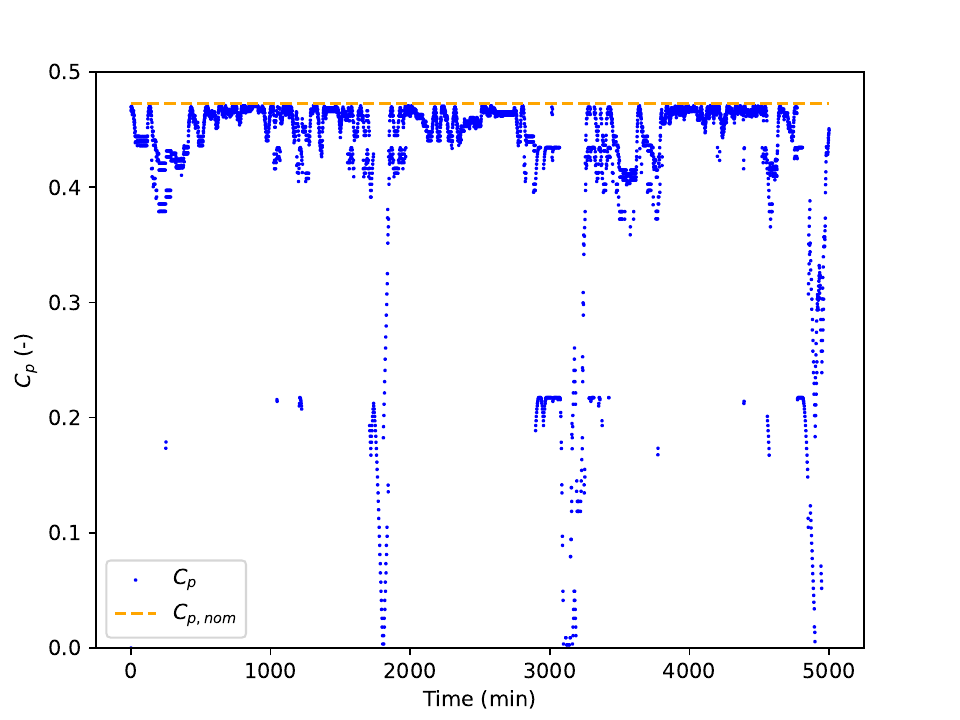}
    \captionsetup{labelfont={bf}}
    \caption{$C_p$ control plot.}
    \label{fig: VI cp exp control}
  \end{subfigure}

  \captionsetup{labelfont={bf}}
  \caption{Controlled variables for the VI agent on the \textit{experimental} environment.}
  \label{fig: VI exp control}
\end{figure}

\color{rev1}
Figure \ref{fig: VI exp control} summarizes the results for the Value Iteration (VI) agent under the \textit{experimental} environment. The control strategy employed by the VI agent differs from that of the DDQN1 agent. The VI agent places greater emphasis on rotor speed control, allowing the yaw misalignment. In Figure \ref{fig: VI yaw exp control}, it is evident that the turbine yaw exhibits a short delay with respect to the wind direction. Despite this misalignment, the intensive rotor speed control enables the VI agent to maintain a more stable value of the tip speed ratio, as depicted in Figure \ref{fig: VI tsr exp control}. Additionally, also Figure \ref{fig: VI tsr exp control} reveals that the VI agent utilizes pitch control with a lower frequency than the DDQN1 agent, as it prioritizes the rotor speed control.

The VI agent achieves a control capacity factor (CCF) of around 90\% but falls short behind the DDQN1 agent (93\%). This discrepancy can be attributed to the sensitivity of the VI algorithm to the model definition. In this case, the VI agent was trained under the assumption that the wind remains constant between actions (see \ref{appendix_Value_Iteration}), which is not the case in the \textit{experimental} wind, where the wind changes at each iteration. This limitation underlines one of the main drawbacks of the VI algorithm, which is to require a precise model of the environment. Certainly, this assumption is virtually true for the \textit{narrow} and \textit{wide} environments, as the frequency of oscillations in the wind conditions is negligible (compared to the number of iterations), that is why the VI achieves almost perfect control on those scenarios, as demonstrated in Figure \ref{fig:performance review}. In a rapidly changing (and unpredictable) environment, DDQN RL control is more appropriate. 
\color{black}

\begin{figure}[H]
  \centering
  \begin{subfigure}{0.49\textwidth}
    \centering
    \includegraphics[width=0.9\linewidth]{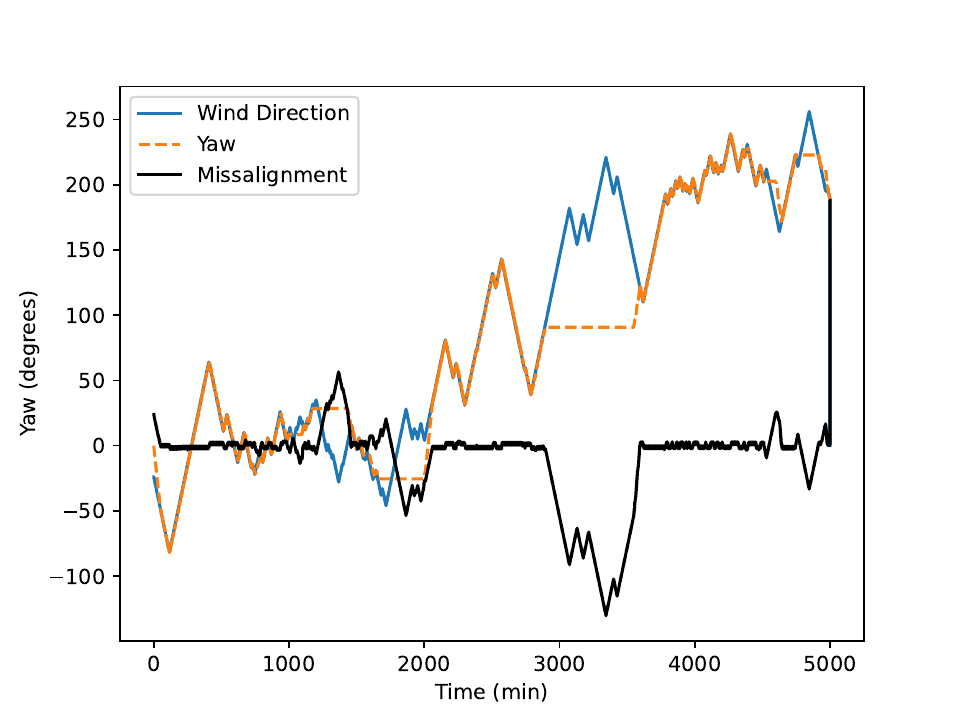}
    \captionsetup{labelfont={bf}}
    \caption{Yaw control plot.}
    \label{fig: yaw exp control PID}
  \end{subfigure}
  \hfill
  \begin{subfigure}{0.49\textwidth}
    \centering
    \includegraphics[width=0.9\linewidth]{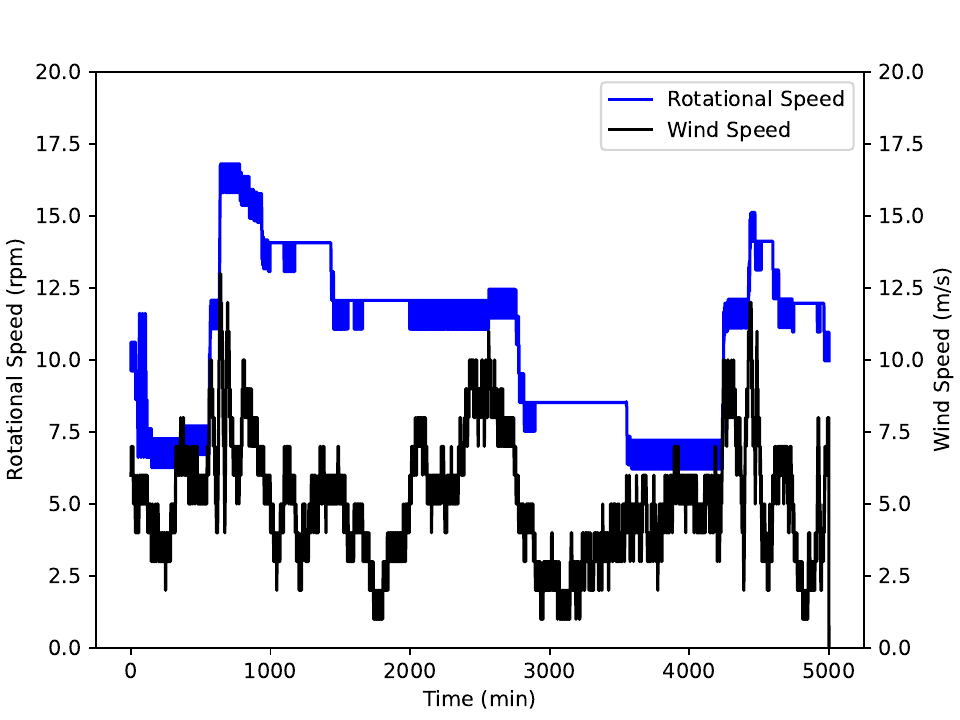}
    \captionsetup{labelfont={bf}}
    \caption{Rotor speed and wind speed control plot.}
    \label{fig:tsr exp control PID}
  \end{subfigure}
  \begin{subfigure}{0.49\textwidth}
    \centering
    \includegraphics[width=0.9\linewidth]{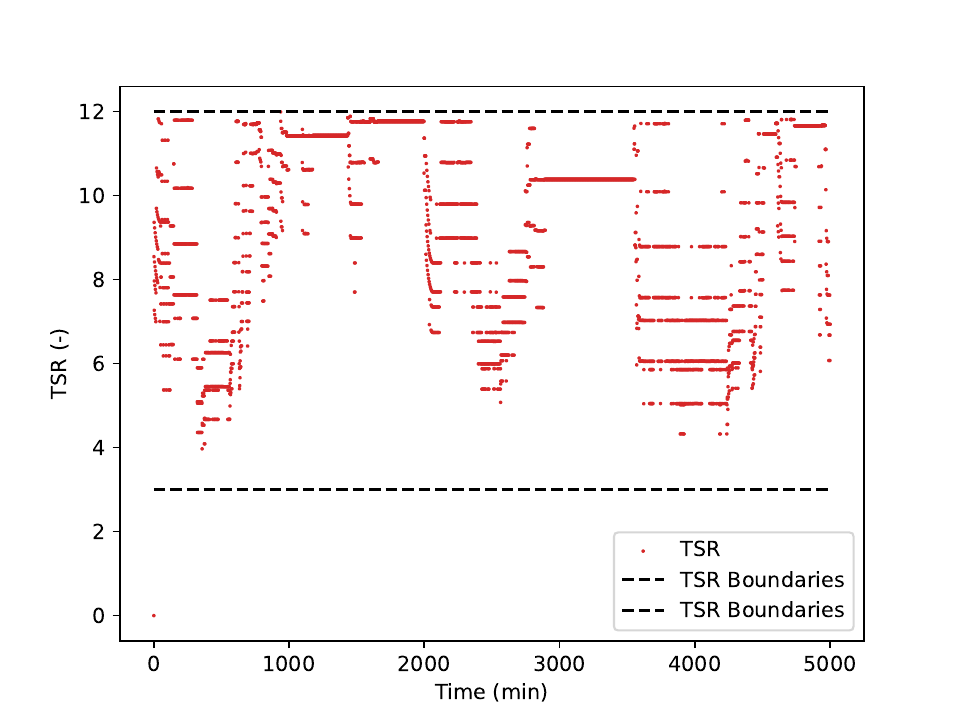}
    \captionsetup{labelfont={bf}}
    \caption{TSR control plot.}
    \label{fig:tsr control PID}
  \end{subfigure}
  \hfill
  \begin{subfigure}{0.49\textwidth}
    \centering
    \includegraphics[width=0.9\linewidth]{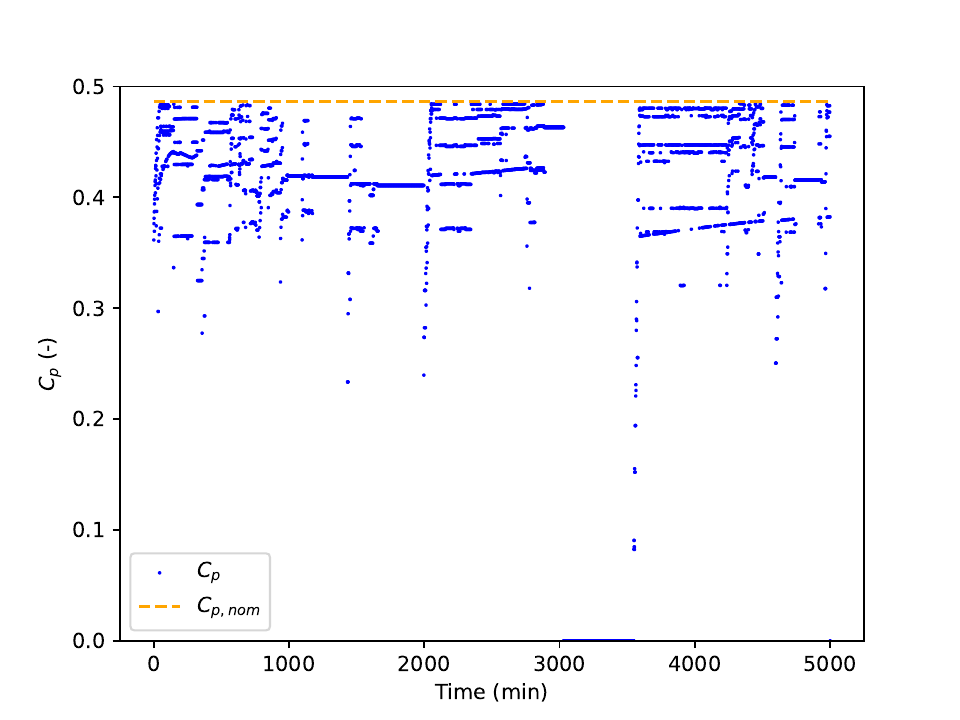}
    \captionsetup{labelfont={bf}}
    \caption{$C_p$ control plot.}
    \label{fig:cp exp PID control}
  \end{subfigure}

  \captionsetup{labelfont={bf}}
  \caption{Controlled variables for the PID agent on the \textit{experimental} environment.}
  \label{fig:exp pid control}
\end{figure}

DDQN and VI results improve those obtained by the classic PID control (see \ref{appendix_PID} for details). The results achieved by the PID control are plotted in Figure \ref{fig:exp pid control}. The main difficulty for the PID control is that it is unable to learn which are the constraints of the state space $S$ and, therefore, often performs forbidden actions. These actions are revoked and, if the agent keeps insisting on doing them, it gets stuck. This problem is not encountered by the RL agents, which only performed forbidden actions for a very small number of times during the training. This can be observed in the second plot: at 3000 time steps, the agent gets stuck on the upper TSR limit (TSR=12) and its actions, which lead to surpassing the limit, keep getting revoked. This impedes the agent from maintaining good alignment with the incoming wind and its misalignment grows to values surpassing 70 degrees during which the agent achieves a $C_p$ of zero. The RL's ability to properly learn the state space is its biggest advantage over the PID control. We have checked that even when the PID control can act three times per time step (the RL agent can only act only once), it fails to perform as well as the RL agent, achieving only a CCF of 76\% while the RL agent achieved 93.3\% in the \textit{experimental} scenario.

\color{rev2}
\subsection{Wind turbine control in 4-months long real wind conditions}
\begin{figure}[h]
  \centering

  \begin{subfigure}{0.49\textwidth}
    \centering
    \includegraphics[width=\textwidth]{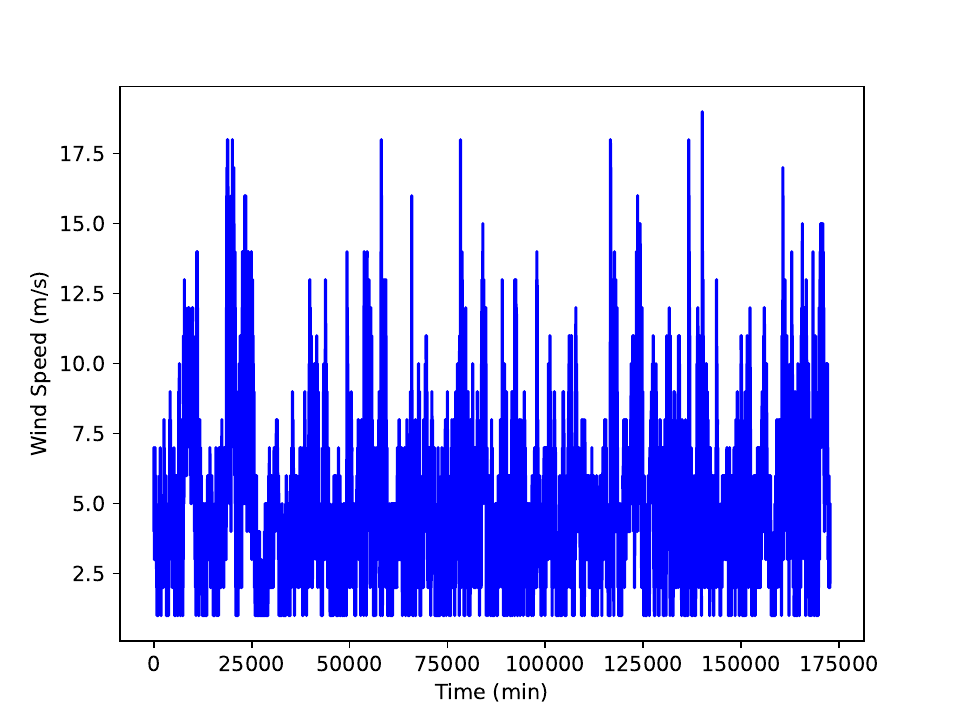}
    \captionsetup{labelfont={bf}}
    \caption{Wind Speed during the four-month scenario.}
    \label{fig:4month_windspeed}
  \end{subfigure}
  \hfill
  \begin{subfigure}{0.49\textwidth}
    \centering
    \includegraphics[width=\textwidth]{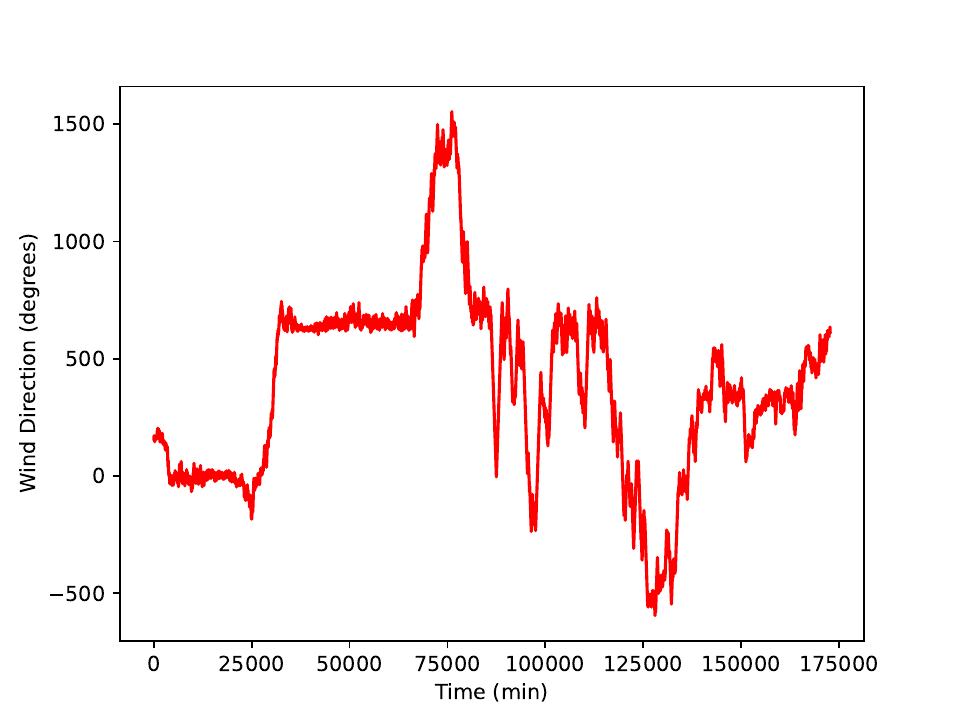}
    \captionsetup{labelfont={bf}}
    \caption{Wind direction during the four-month scenario.}
    \label{fig:4month_winddir}
  \end{subfigure}
  
  \captionsetup{labelfont={bf}}
  \caption{Wind speed and direction during the four-month scenario.}
  \label{fig:4months_experimental_wind}
\end{figure}

The previous results prove that the agent is capable of controlling a turbine in short but realistic scenarios. In light of the previous success, a final scenario is proposed to challenge the RL framework with 4-month real unsteady-turbulent wind.
With this new scenario, we can compute performance metrics such as yearly production or control capacity factors. Moreover, this scenario is intended to showcase the ability of the different strategies to control the turbine over a long time with more variations in wind conditions. We chose the real wind from Spring 2023 sampled from \cite{JAA96} for four months. The wind speed and wind direction are shown in Figure \ref{fig:4month_windspeed}
 and \ref{fig:4month_winddir}. 

\begin{figure}[H]
  \centering
  \begin{subfigure}{0.49\textwidth}
    \centering
    \includegraphics[width=0.9\linewidth]{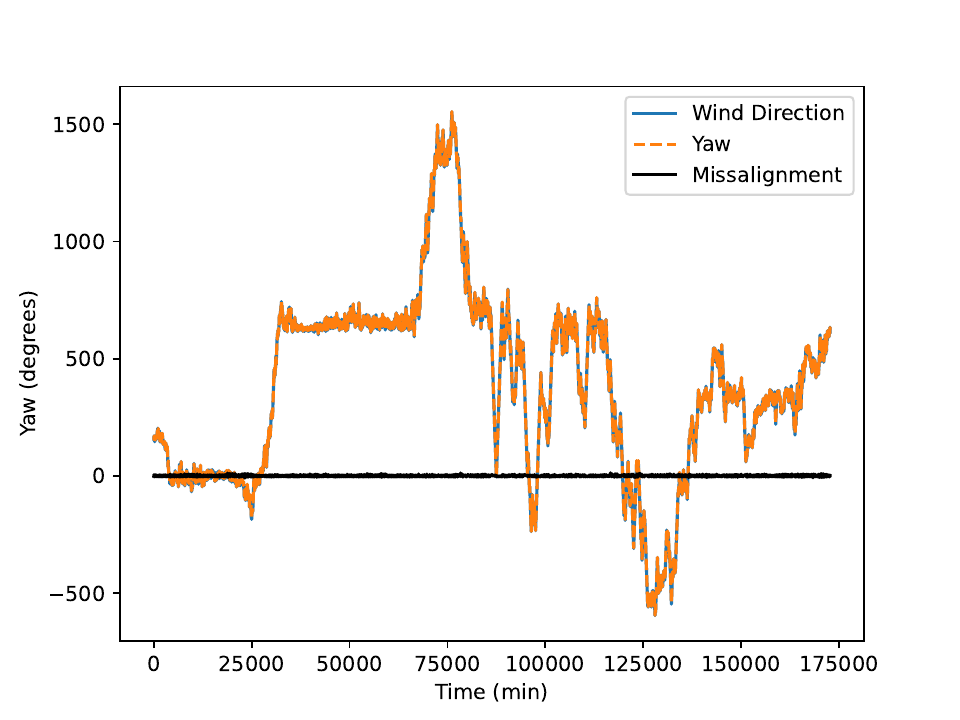}
    \captionsetup{labelfont={bf}}
    \caption{Yaw control plot.}
    \label{fig: 4month yaw control}
  \end{subfigure}
  \hfill
  \begin{subfigure}{0.49\textwidth}
    \centering
    \includegraphics[width=0.9\linewidth]{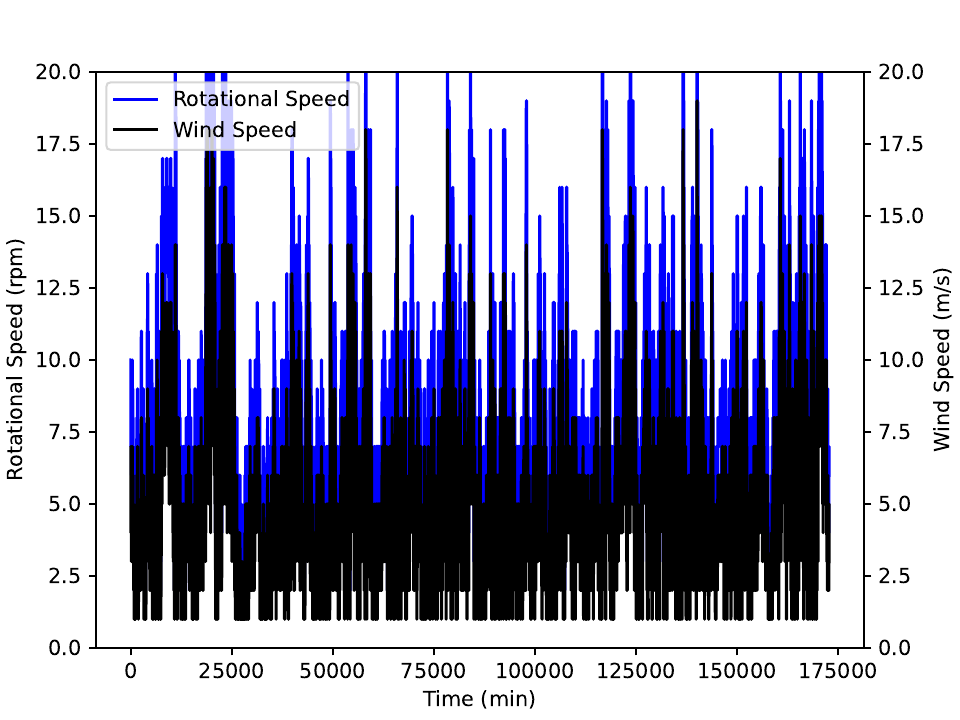}
    \captionsetup{labelfont={bf}}
    \caption{Rotor speed and wind speed control plot.}
    \label{fig:4month rpm control}
  \end{subfigure}

  \begin{subfigure}{0.49\textwidth}
    \centering
    \includegraphics[width=0.9\linewidth]{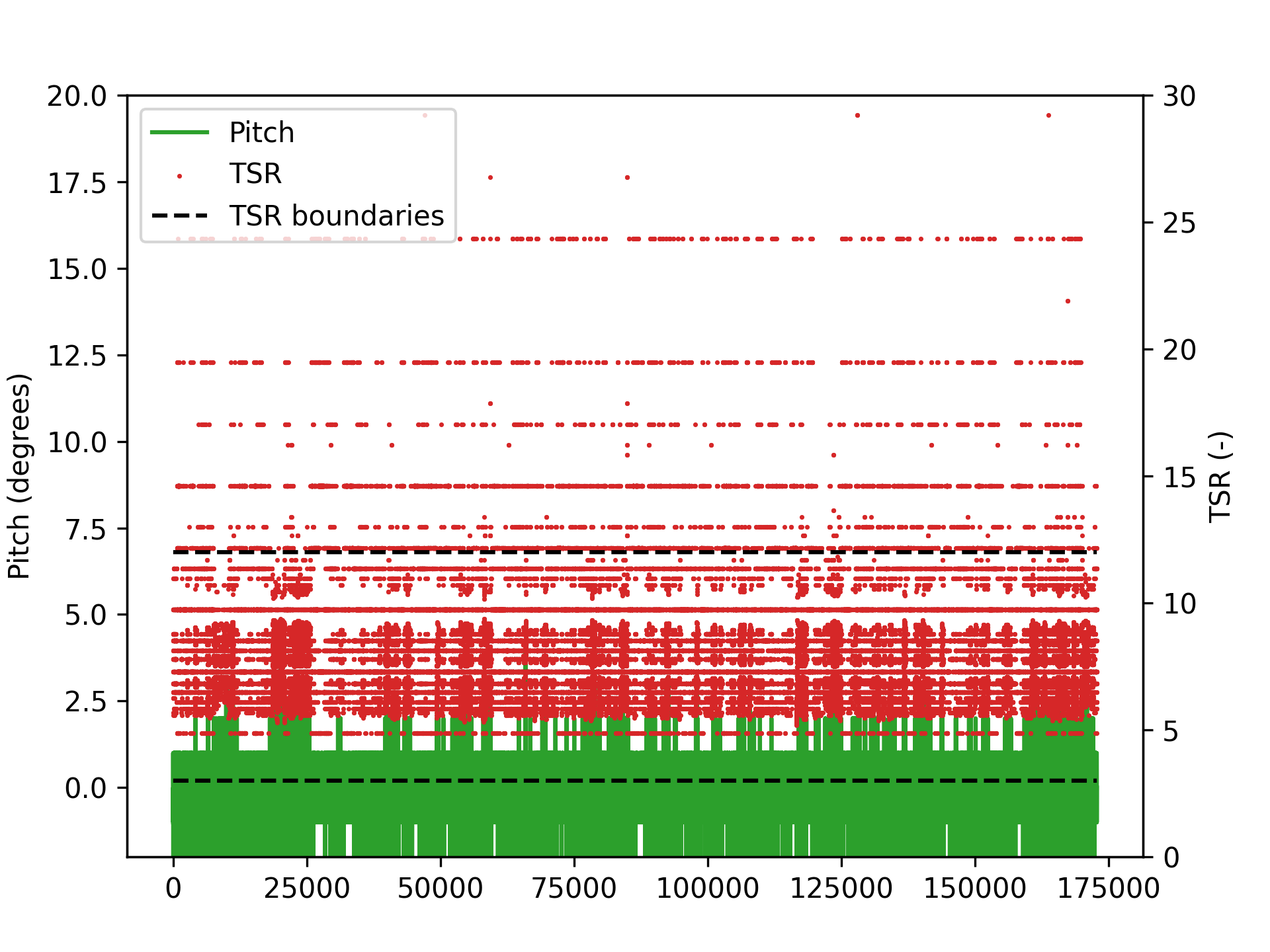}
    \captionsetup{labelfont={bf}}
    \caption{TSR evolution and pitch control plot.}
    \label{fig:4month tsr control}
  \end{subfigure}
  \hfill
  \begin{subfigure}{0.49\textwidth}
    \centering
    \includegraphics[width=0.9\linewidth]{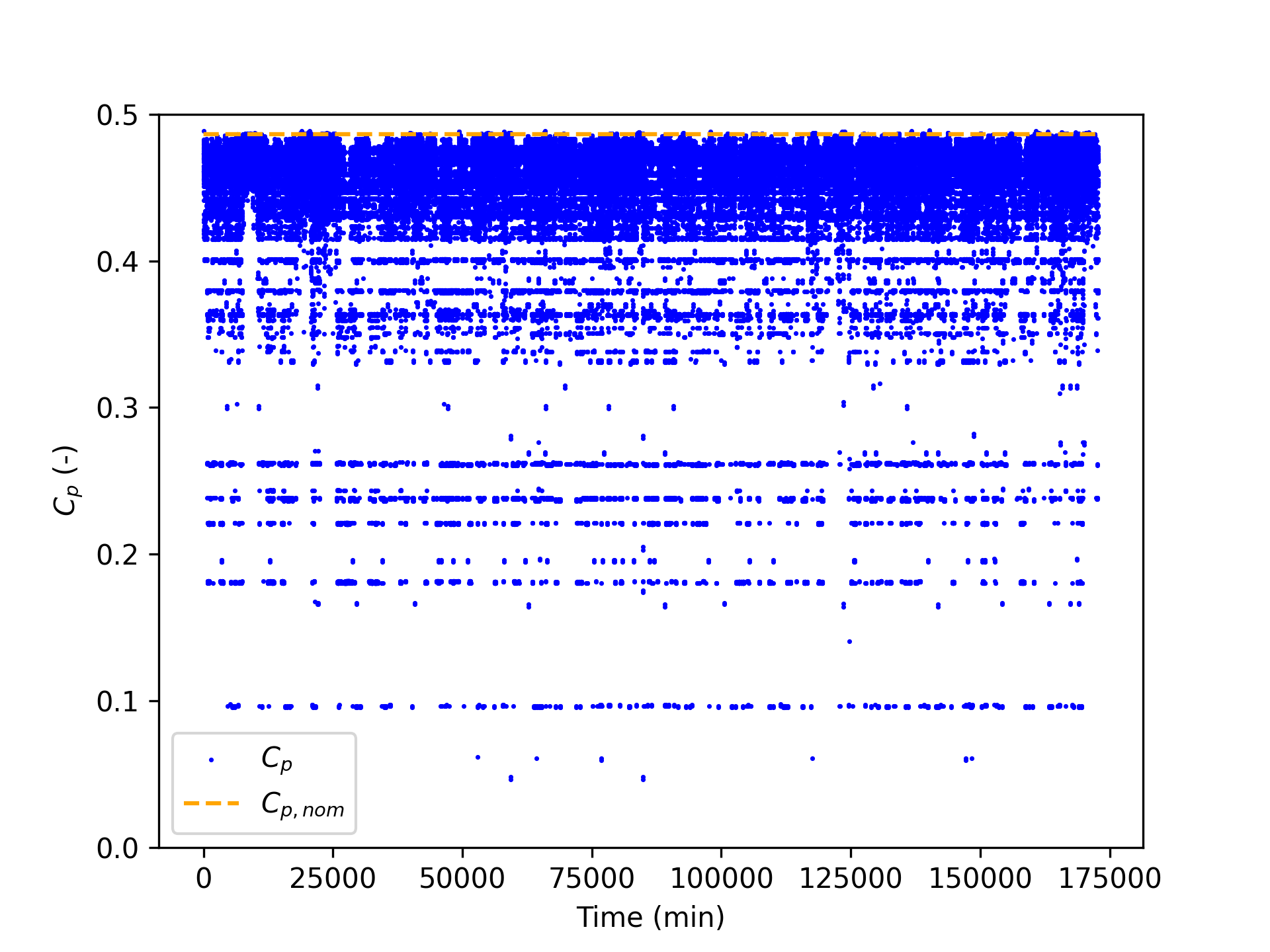}
    \captionsetup{labelfont={bf}}
    \caption{$C_p$ control plot.}
    \label{fig:4months cp exp control}
  \end{subfigure}

  \captionsetup{labelfont={bf}}
  \caption{Controlled variables for the optimized RL agent on the four-month scenario.}
  \label{fig:4month control}
\end{figure}

The results of the DDQN1 RL agent are shown in Figure \ref{fig:4month control}. In this setting, the agent proves again to be able to maintain perfect alignment whilst maintaining optimal aerodynamic efficiency of its blades through carefully control of the rotor speed and the pitch angle. Again, the TSR exceeds the specified limits due to, as explained earlier, lack of exposure to wind velocities below the cut-in speed during training. Despite this shortcoming, the agent is able to adapt to these situations well. Furthermore, even though the wind speeds are below the cut-in speeds in 29.7\% of iterations and are thus unknown to the agent, it only commands a TSR outside the allowed range in 8.4\% of iterations. Therefore, the agent displays its generalization capabilities by adequately controlling a large volume of unseen states, only committing errors at a low rate. This control yields a high average $C_p$, with a control capacity factor of 91.39\%.

Finally, Table \ref{tab:4month metrics comparison} compares the different controls, for the 4-month environment. The results include the control capacity factor ($CCF = \frac{\langle C_p\rangle}{C_{p,nom}} \cdot 100\%$), the capacity factor ($CF = \frac{E}{P_{nom}\cdot T}$) and the total energy production (scaled from 4 months to a full year).
Again, the reinforcement learning agents outperform the classic PID control, with the DDQN emerging as the most effective. Furthermore, these results showcase the significance of a robust control strategy capable of adapting the operational parameters of the wind turbine to varying wind conditions, enabling the wind plant to maximize energy capture.

\begin{table}[htbp]
\centering
\begin{tabular}{l||c|c|c|c}
\hline
\bf Metric & \bf DDQN1 & \bf VI & \bf PID & \bf Uncontrolled \\ \hline
Control Capacity Factor (\%) & 91.31 & 87.50 & 57.60 & 12.77 \\
Capacity Factor (\%) & 20.95 & 20.50 & 12.49 & 1.59\\
Yearly Production (MWh) & 4162.95 & 4073.45 & 2481.97 & 316.12 \\ \hline
\end{tabular}
\captionsetup{labelfont={bf}}
\caption{Performance metrics (Control Capacity Factor, Capacity Factor and Yearly Production) for different agents on the four-months scenario.}
\label{tab:4month metrics comparison}
\end{table}

\color{black}
\section{Conclusions}

Reinforcement learning has emerged as a promising approach to wind turbine control, offering the potential to optimize performance in different environments while showing better behavior and flexibility than classic controls, such as PID. By leveraging RL algorithms such as Q-learning, wind turbine controllers can adapt and learn optimal actions in real time, considering the dynamic and uncertain nature of the wind environment. Through the iterative learning process, RL enables wind turbines to make informed decisions, improve energy capture efficiency, and enhance turbine reliability in changing environments.

RL algorithms can optimize various control parameters, including pitch angle, rotor speed, and yaw angle, to maximize power output. By autonomously learning from operational data, RL-based controllers can adapt to changing wind conditions and optimize turbine operation under various scenarios. \textcolor{rev2}{In this work, we have shown that double
deep Q-learning reinforcement learning outperforms PID control and classic value iteration reinforcement learning when maximizing wind turbine energy generation in changing environments, including real turbulent/gusty winds.}

Future research directions include exploring hybrid control approaches that combine RL with traditional control strategies, leveraging the strengths of both methods. Moreover, investigating the potential of multi-agent RL algorithms for cooperative control of wind turbine arrays could further enhance overall system performance and grid integration. 

\appendix 

\section{Value Iteration Reinforcement Learning}\label{appendix_Value_Iteration}
\color{rev1}
Value Iteration (VI) \cite{sutton1998rli} stands as a classic model-based RL algorithm designed to compute the optimal policy within the framework of a Markov Decision Process (MDP). Value Iteration is a Dynamic Programming algorithm which provides an efficient way to calculate these value functions. It is based on the original Bellman equation, a fundamental element in every RL algorithm. The pseudocode used for the VI algorithm is presented in Algorithm \ref{alg:Value_Iteration}. VI algorithms are efficient but come with two primary drawbacks: firstly, being model-based, they need an exact model of the environment, a condition often unattainable in practical scenarios; secondly, these algorithms exhibit poor scalability, with computational costs escalating significantly when the state space grows. Despite these limitations, Value Iteration remains a valuable reference algorithm, providing a straightforward means to obtain the optimal behavior when the state space is finite and small enough.

The behavior of the environment is given by a set of probabilities $p(s',r|s,a)$ that determines the probability of reaching a state $s'$ and to receive a reward $r$, given that you are in the state $s$ and perform the action $a$.
To discriminate between desirable and undesirable states, \textit{value functions}, $V(s)$, are used. The value function provides the maximum cumulative reward, taking into account the long term, achievable from a specific state $s$. Therefore, the optimal policy, $\pi^*(s)$, is defined by the action that allows the agent to maximize its value function for the next time step, that is $\pi^*(s)=\text{argmax}_a V(s')$, where $s'$ denotes the state we reach after we take action $a$ in the state $s$. 

To adapt this methodology to wind turbine control, both the state and action space need to be defined discretely, following a similar approach to the DDQN agent, as detailed in Section \ref{subsec:RL_strategy}. Additionally, the transition probability function, $p(s'|s,a)$, must be defined. For simplicity, we assume that wind conditions do not change between actions, making the transition function deterministic. The computational cost of the VI is determined by  the calculation of the reward function for every state. This involves solving the wind turbine model for every possible state, emphasizing the importance of employing an efficient and rapid solver, such as the Blade Element Momentum Theory (BEMT), in this work.

\color{black}
\begin{algorithm}[H]
    \caption{Value Iteration \cite{sutton1998rli}}
    \label{alg:Value_Iteration}
    \SetAlgoLined
    \SetKwInOut{Parameter}{Algorithm parameter}
    \Parameter{$\varepsilon > 0$ determining accuracy of estimation}
    Initialize $V(s)$, for all $s \in S^+$, arbitrarily except that $V(\text{terminal}) = 0$\;
    \While{$\Delta \geq \varepsilon$}{
        $\Delta \gets 0$\;
        \For{each $s \in S$}{
            $v \gets V(s)$\;
            $V(s) \gets \displaystyle\max_a \displaystyle\sum_{s', r} p(s', r | s, a) [r + \gamma V(s')]$\;
            $\Delta \gets \max(\Delta, |v - V(s)|)$\;
        }
    }
    \For{each $s \in S$}{
        $\pi(s) \gets \displaystyle\arg\max_a \displaystyle\sum_{s', r} p(s', r | s, a) [r + \gamma V(s')]$\;
    }
    \KwOut{Deterministic policy $\pi$}
\end{algorithm}

\section{PID design}\label{appendix_PID}
\begin{figure}[H]
  \centering
  \includegraphics[scale=0.5]{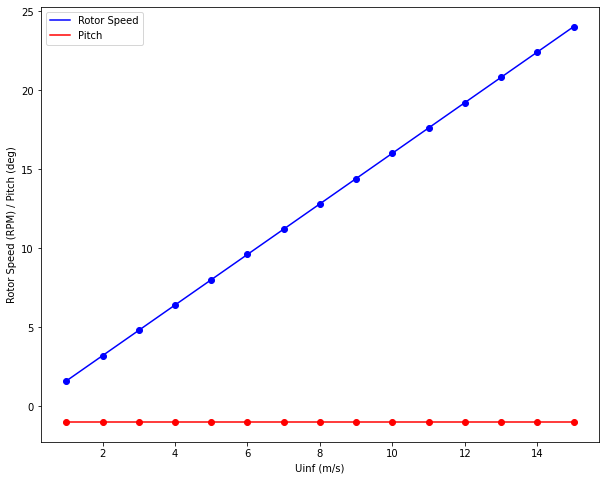}
  \captionsetup{labelfont={bf}}
  \caption{Pitch and rotor speed PID setpoints.}
  \label{fig: setpoints}
\end{figure}

The classic PID control has been designed to mimic the RL agent, to provide a fair comparison. The PID has changing setpoints that allow it to adapt to changing winds and maintain optimal working conditions. These setpoints are designed to make the agent follow the same strategy as the RL agent, that is, controlling efficiency through the rotor speed rather than the pitch angle. With that objective in mind, the setpoints presented in Figure \ref{fig: setpoints} are implemented. If the agent is able to follow them, the turbine can work at the optimum TSR, and, if paired with a yaw misalignment equal to 0, it will achieve perfect control.

The PID control is unable to learn the limitations of the state space, but it still has to abide by them.  When the PID performs a forbidden action, the action is revoked in the same way as it is revoked for the RL agent, that is, if $s_{t+1} \notin S$ then $s_{t+1} = s_t$. The PID lacks the ability to learn the boundaries of $S$, which is the main limiting factor for its performance. It is also relevant to note that the PID agent is, in fact, composed of three different PID controllers, each acting on a different control variable. This means that the agent is acting three times per turn, much more than the single action per turn of the RL agent.
Moreover, to make the PID control comparable to the RL agent, the PID actions are limited to be +1,-1 or 0 for each control variable. This agent is implemented using the Simple PID library \cite{L23} and the agent's gains (proportional, derivative, and integral) are tuned using a Bayesian search to improve its performance.
	
\section*{Acknowledgments}
Esteban Ferrer and Oscar Mariño would like to thank the support of
Agencia Estatal de Investigación for the grant "Europa Excelencia" for the project EUR2022-134041 funded by MCIN/AEI/10.13039/501100011033) y the European Union NextGenerationEU/PRTR. Esteban Ferrer and Martin de Frutos also acknowledge the help of the Comunidad de Madrid and Universidad Politécnica de Madrid for the Young Investigator award: APOYO-JOVENES-21-53NYUB-19-RRX1A0. 
This research has been cofunded by the European Union (ERC, Off-coustics, project number 101086075). Views and opinions expressed are, however, those of the author(s) only and do not necessarily reflect those of the European Union or the European Research Council. Neither the European Union nor the granting authority can be held responsible for them.
Finally, all authors gratefully acknowledge the Universidad Politécnica de Madrid (www.upm.es) for providing computing resources on the Magerit Supercomputer.

\bibliographystyle{abbrv}
\bibliography{biblio}

\end{document}

%% file: NN_scheme.tex
\begin{tikzpicture}[scale=0.75,node distance=1.5cm]
    \node[circle, fill=green!20,draw=green!60, line width=1.5pt, minimum size=0.5cm] (input-1) at (0,-1) {$s_1$};
    \node[circle, fill=green!20,draw=green!60, line width=1.5pt, minimum size=0.5cm] (input-2) at (0,-3) {$s_4$};
    \node[circle, fill=green!20, draw=green!60, line width=1.5pt, minimum size=0.5cm] (input-3) at (0,-4) {$a_1$};
    \node[circle, fill=green!20, draw=green!60, line width=1.5pt, minimum size=0.5cm] (input-4) at (0,-6) {$a_7$};
    
    \node[circle, fill=blue!20,draw=blue!60, line width=1.5pt, minimum size=0.5cm] (hidden-1) at (2,-2) {\tiny{ReLu}};
    \node[circle, fill=blue!20,draw=blue!60, line width=1.5pt, minimum size=0.5cm] (hidden-2) at (2,-3) {\tiny{ReLu}};
    \node[circle, fill=blue!20,draw=blue!60, line width=1.5pt, minimum size=0.5cm] (hidden-3) at (2,-5) {\tiny{ReLu}};
    
    \node at (0,-1.85) {$\vdots$};
    \node at (0,-4.85) {$\vdots$};
    
    \node[circle, fill=blue!20, draw=blue!60, line width=1.5pt, minimum size=0.5cm] (hidden2-1) at (4,-2) {\tiny{ReLu}};
    \node[circle, fill=blue!20, draw=blue!60, line width=1.5pt, minimum size=0.5cm] (hidden2-2) at (4,-3) {\tiny{ReLu}};
    \node[circle, fill=blue!20, draw=blue!60, line width=1.5pt, minimum size=0.5cm] (hidden2-3) at (4,-5) {\tiny{ReLu}};
    
    \node[circle, fill=blue!20, draw=blue!60, line width=1.5pt, minimum size=0.5cm] (hidden3-1) at (6,-2) {\tiny{ReLu}};
    \node[circle, fill=blue!20, draw=blue!60, line width=1.5pt, minimum size=0.5cm] (hidden3-2) at (6,-3) {\tiny{ReLu}};
    \node[circle, fill=blue!20, draw=blue!60, line width=1.5pt, minimum size=0.5cm] (hidden3-3) at (6,-5) {\tiny{ReLu}};
    
    \node at (2,-3.85) {$\vdots$};
    \node at (4,-3.85) {$\vdots$};
    \node at (6,-3.85) {$\vdots$}; 
    
    \node[circle, fill=red!20, draw=red!60, line width=1.5pt, minimum size=0.5cm] (output) at (8,-3.5) {$Q$};
    
    \foreach \i in {1,2,3,4}
    \foreach \j in {1,2,3}
    \draw[->] (input-\i) -- (hidden-\j);
    
    \foreach \i in {1,2,3}
    \foreach \j in {1,2,3}
    \draw[->] (hidden-\i) -- (hidden2-\j);
    
    \foreach \i in {1,2,3}
    \foreach \j in {1,2,3}
    \draw[->] (hidden2-\i) -- (hidden3-\j);
    
    \foreach \i in {1,2,3}
    \draw[->] (hidden3-\i) -- (output);
    
    \node[above=0.5cm, align=center, font=\scriptsize] at (input-1.north) {Input layer};
    \node[above=0.5cm, align=center, font=\scriptsize] at (hidden2-1.north) {Hidden layers};
    \node[above=0.5cm, align=center, font=\scriptsize] at (output.north) {Output neuron};
    \node[below=0.5cm, align=center, font=\scriptsize] at (hidden-3.south) {256 neurons};
    \node[below=0.5cm, align=center, font=\scriptsize] at (hidden2-3.south) {128 neurons};
    \node[below=0.5cm, align=center, font=\scriptsize] at (hidden3-3.south) {64 neurons};
\end{tikzpicture}